\documentclass[11pt,letterpaper]{article}
\usepackage[OT1]{fontenc}
\usepackage[margin=1in]{geometry}
\usepackage{amsmath,amssymb,amsthm}
\theoremstyle{plain}
\newtheorem{theorem}{Theorem}[section]
\newtheorem{lemma}[theorem]{Lemma}
\newtheorem{proposition}[theorem]{Proposition}
\newtheorem{corollary}[theorem]{Corollary}
\theoremstyle{definition}
\newtheorem{assumption}[theorem]{Assumption}
\theoremstyle{remark}
\newtheorem{remark}[theorem]{Remark}
\usepackage{smile}
\usepackage{mathtools}
\IfFileExists{enumitem.sty}{\usepackage{enumitem}}{}
\usepackage{tabularx}
\usepackage{float}
\usepackage[protrusion=true,expansion=true]{microtype}
\usepackage{xcolor}
\definecolor{linkcolor}{HTML}{ED1C24}
\definecolor{tsinghuapurple}{RGB}{102,8,116}
\usepackage[colorlinks=true,linkcolor=linkcolor,anchorcolor=blue,citecolor=blue,urlcolor=black]{hyperref}
\setcitestyle{authoryear,round,semicolon}

\allowdisplaybreaks
\newcommand{\EffectiveBias}{\mathrm{EffectiveBias}}
\newcommand{\EffectiveVar}{\mathrm{EffectiveVar}}
\newcommand{\bias}{\mathrm{bias}}
\newcommand{\variance}{\mathrm{variance}}
\newcommand{\noise}{\mathrm{noise}}
\newcommand{\drift}{\mathrm{drift}}
\newcommand{\burn}{\mathrm{burn}}
\newcommand{\wt}[1]{\widetilde{#1}}

\newcommand{\thupurple}[1]{{\color{tsinghuapurple}#1}}

\title{\huge Settling the Burn-in Gap: Sharp Risk Bounds for Tail-Averaged SGD in Linear Regression}

\author
{
    Qingyue Zhao\thanks{Equal contribution} \thanks{Department of Computer Science, University of California, Los Angeles, CA 90095, USA; e-mail: {\tt zhaoqy24@cs.ucla.edu}}
    ~~
    Xuheng Li\footnotemark[1] \thanks{Department of Computer Science, University of California, Los Angeles, CA 90095, USA; e-mail: {\tt xuheng.li@cs.ucla.edu}} 
    ~~~~
    Quanquan Gu\thanks{Department of Computer Science, University of California, Los Angeles, CA 90095, USA; e-mail: {\tt qgu@cs.ucla.edu}}
}

\date{}
\begin{document}
\maketitle
\begin{abstract}
We study the excess risk of constant-stepsize online stochastic gradient descent (SGD) with tail-averaging for high-dimensional linear regression. Building on the instance-dependent spectral framework of \citet{zou2021}, we establish matching initialization-explicit upper and lower bounds on the total excess risk of tail-averaged SGD in terms of the population covariance spectrum $\Hb$ and initialization $\wb_0-\wb^*$ whenever a trajectory spectral feedback condition holds. Existing analyses exhibit a fundamental gap in the effective variance: the upper bound of \citet{zou2021} retains an undamped contribution from the initial error in the head eigenspace $\Ib_{0:k^\dagger}$, whereas their lower bound omits the variance induced by the head initialization altogether. We show that both the existing upper and lower bounds are loose: the single-hop fast-mode variance contribution from the head initialization is exponentially damped during the burn-in phase, whereas fourth-moment energy scattered from the head initialization into slow spectral modes during the early burn-in steps survives undamped into the averaging window. We prove that our spectral feedback condition is satisfied in two concrete constant-stepsize regimes: (i) for arbitrary covariance spectra $\Hb$ and any finite horizon, under an explicit horizon-dependent stepsize chosen prior to execution; and (ii) for polynomially decaying spectra $\lambda_i\asymp i^{-a}$ ($a>1$), under a single horizon-independent positive stepsize. Finally, we complement these positive results with explicit one- and two-dimensional separation constructions proving an exponential-versus-polynomial gap in the lower bound of \citet{zou2021} and showing why fourth-moment assumptions alone cannot determine the asymptotic exponential decay rate without additional structural conditions.\\
Proof ideas emerged from interactive brainstorming between the authors and the AI models GPT-6 Astra and Gemini 4 Argon; the authors simplified, reorganized, and verified all proofs and take full responsibility for the paper.
\end{abstract}

\section{Introduction}\label{sec:1}
Understanding how first-order stochastic optimization generalizes in high- and infinite-dimensional feature spaces is a central theme in modern statistical learning theory. For offline empirical risk minimization in linear regression, \citet{bartlett2020} and \citet{tsigler2023} characterized \emph{benign overfitting}---when minimum-$\ell_2$-norm interpolators and ridge regression generalize well despite fitting noisy training data---in terms of the effective ranks of the population data covariance operator $\Hb$ \citep{belkin2020,hastie2022}. In the complementary setting of fresh-sample online stochastic approximation, \citet{zou2021} established an instance-dependent excess risk analysis for constant-stepsize stochastic gradient descent (SGD) with full iterate averaging, demonstrating that online SGD enjoys an implicit spectral regularization effect governed by a closely related spectral partition of $\Hb$ (while noting that for polynomially decaying spectra such as $\lambda_i=i^{-2}$, online SGD achieves vanishing excess risk by forgetting noisy labels rather than interpolating training data; see also \citet{wu2025}).

In large-scale stochastic optimization, however, averaging all iterates from the initial point $\wb_0$ suffers from slow polynomial decay of its deterministic drift bias $\frac12\norm{\EE[\overbar{\wb}_{0:N}]-\wb^*}_{\Hb}^2$---decaying at best as $\cO(N^{-2})$ when $\norm{\wb_0-\wb^*}_{\Hb^{-1}}<\infty$ (or in finite dimensions with $\lambda_d>0$), and even more slowly (e.g., $\Theta(N^{-3/2})$ when $\lambda_i=i^{-2}$ and $(\wb_0-\wb^*)_i^2=i^{-2}$) when slow spectral modes remain uncontracted. To accelerate the decay of initial error while preserving statistical efficiency, \citet{rakhlin2012} introduced \emph{suffix (tail) averaging} for stochastic gradient descent, and \citet{jain2017b,jain2018} developed its operator-theoretic analysis for least-squares linear regression; in modern deep learning and language model pretraining, its checkpoint-based counterpart \emph{Latest Weight Averaging (LAWA)} \citep{kaddour2022,sanyal2023} averages recent checkpoints along the trajectory to achieve rapid training speedups. Tail-averaging discards the first $s$ iterates as a burn-in phase and averages the subsequent $N$ iterates:
\[
\overbar{\wb}_{s:s+N}=\frac1N\sum_{t=s}^{s+N-1}\wb_t,
\qquad\text{where}\qquad
\wb_t=\wb_{t-1}+\gamma\bigl(y_t-\la\wb_{t-1},\xb_t\ra\bigr)\xb_t.
\]
\citet[Theorems~5.1 and~5.2]{zou2021} extended their spectral excess risk analysis to tail-averaged SGD. Let $\{\lambda_i\}_{i\ge1}$ denote the eigenvalues of $\Hb$ in non-increasing order, and define the effective dimension cutoffs $k^*=\max\{k:\lambda_k\ge(\gamma N)^{-1}\}$ and $k^\dagger=\max\{k:\lambda_k\ge(\gamma(s+N))^{-1}\}$. Up to universal multiplicative constants, when the burn-in length satisfies $s=\Theta(N)$, \citet[Theorem~5.1]{zou2021} upper-bounds the bias-induced variance contribution by
\begin{equation}\label{eq:1.1}
\frac{\alpha\bigl(\norm{\wb_0-\wb^*}_{\Ib_{0:k^\dagger}}^2+(s+N)\gamma\norm{\wb_0-\wb^*}_{\Hb_{k^\dagger:\infty}}^2\bigr)}{N\gamma(1-\gamma\alpha\tr(\Hb))}
\cdot\left(\frac{k^*}{N}+N\gamma^2\sum_{i>k^*}\lambda_i^2\right),
\end{equation}
whereas their lower bound \citep[Theorem~5.2]{zou2021} yields only
\begin{equation}\label{eq:1.2}
\beta\norm{\wb_0-\wb^*}_{\Hb_{k^\dagger:\infty}}^2
\cdot\left(\frac{k^*}{N}+N\gamma^2\sum_{i>k^*}\lambda_i^2\right).
\end{equation}
Comparing \eqref{eq:1.1} and \eqref{eq:1.2}, \citet{zou2021} observed that the two bounds fail to match whenever the initial error $\wb_0-\wb^*$ has significant energy in the head subspace $\Ib_{0:k^\dagger}$, and conjectured that the upper bound \eqref{eq:1.1} can be improved. The same discrepancy on the head initialization subspace persists in the exponential moving average (EMA) analysis of \citet[Theorems~4.1 and~4.2]{li2025} and in the simplified diagonal covariance analysis of \citet{meterez2025}.

\paragraph{Why Both Existing Upper and Lower Bounds Are Loose.}
In this paper, we show that closing the gap between \eqref{eq:1.1} and \eqref{eq:1.2} requires sharpening both the upper bound in \citet[Theorem~5.1]{zou2021} and the lower bound in \citet[Theorem~5.2]{zou2021}. Specifically, expanding the outer product in \eqref{eq:1.1} between the initialization subspaces $\{\Ib_{0:k^\dagger},\Hb_{k^\dagger:\infty}\}$ and the fast and slow variance components $\{k^*/N,N\gamma^2\sum_{i>k^*}\lambda_i^2\}$ reveals two distinct physical mechanisms during the burn-in phase:
\begin{enumerate}
\item \emph{Looseness in the Upper Bound (Undamped Fast-Mode Terms):}
The product term $\frac1{\gamma N}\norm{\wb_0-\wb^*}_{\Ib_{0:k^\dagger}}^2\cdot\frac{k^*}{N}$ in \eqref{eq:1.1} does not decay with the burn-in length $s$. Consequently, even when all eigenvalues satisfy $\lambda_i\ge(\gamma N)^{-1}$ ($k^*=d$ and $\Hb_{k^*:\infty}=\zero$), \eqref{eq:1.1} predicts a polynomial rate $\cO(d/(\gamma N^2))$ instead of exponential burn-in decay \citep{jain2018}. This looseness arises because prior upper-bound analyses drop the deterministic burn-in contraction acting on multiplicative gradient fluctuations after they are excited. Retaining the exact burn-in propagation shows that the single-hop $(i\le k^*,j\le k^*)$ contribution is exponentially damped in $s$ through $(\Ib-\gamma\Hb)^s(\wb_0-\wb^*)$ and a burn-in cross-spectral convolution (Theorems~\ref{thm:4.1} and~\ref{thm:4.2}), while multi-hop feedback (including energy that scatters into slow modes $i>k^*$ and subsequently returns to fast modes) is absorbed into the total excess risk whenever our spectral feedback condition $\chi_{s,N}\le\chi_0<1$ holds.

\item \emph{Looseness in the Lower Bound (Burn-In Head-to-Tail Transfer):}
Conversely, whenever slow spectral modes $i>k^*$ exist ($k^*<d$), the cross-interaction between the initial head error term $(\gamma N)^{-1}\norm{\wb_0-\wb^*}_{\Ib_{0:k^\dagger}}^2$ and the slow-spectrum factor $N\gamma^2\sum_{i>k^*}\lambda_i^2$ \emph{cannot} be removed from the total-risk upper bound; instead, it is a genuine lower bound contribution missing from \eqref{eq:1.2}. During the first $\Theta(1/(\gamma\lambda_j))$ steps of the burn-in phase, an initial error component along a fast eigenvector $\vb_j$ ($j\le k^\dagger$) undergoes multiplicative gradient fluctuations $(\xb_t\xb_t^\top-\Hb)(\wb_{t-1}-\wb^*)$. Under Assumption~\ref{ass:3.4}, this process scatters fourth-moment energy proportional to $\beta\gamma\norm{\wb_0-\wb^*}_{\Ib_{0:k^\dagger}}^2\Hb$ across all eigendirections of $\Hb$. While the single-hop energy scattered into fast modes ($i\le k^*$) decays over the remaining burn-in steps, the energy scattered into slow modes ($i>k^\dagger$, where $\lambda_i<(\gamma(s+N))^{-1}$) experiences negligible contraction $(1-\gamma\lambda_i)^{2(s-1-t)}\ge e^{-2}$ and survives undamped into the averaging window---even in two-dimensional noiseless Gaussian regression (Proposition~\ref{prop:4.7}(1)).
\end{enumerate}

\paragraph{Main Contributions.}
Our main contributions are summarized as follows (see Table~\ref{tab:1} in Section~\ref{sec:4.3} for a detailed term-by-term comparison with \citet{zou2021}):
\begin{itemize}
\item \textbf{Matching Excess Risk Bounds:}
Under the distributional assumptions of \citet{zou2021}---Assumptions~\ref{ass:3.1}--\ref{ass:3.3} for the upper bound (Theorem~\ref{thm:4.1}) and the well-specified model \eqref{eq:3.4} with Assumptions~\ref{ass:3.1}, \ref{ass:3.2}, and~\ref{ass:3.4} for the lower bound (Theorem~\ref{thm:4.2})---we derive an exact trajectory operator-polynomial identity (Lemma~\ref{lem:5.1}) and control repeated fourth-moment interactions via a scalar spectral feedback ratio $\chi_{s,N}$ (Lemma~\ref{lem:5.2}). Whenever $\chi_{s,N}\le\chi_0<1$, the multi-hop feedback terms are absorbed within a factor of $(1-\chi_{s,N})^{-1}$, yielding explicit spectral upper and lower bounds for $\EffectiveBias$ and $\EffectiveVar$ that match up to constants depending only on $\alpha$, $\beta$, and $1-\chi_0$ across all dimensions, initializations $\wb_0-\wb^*$, burn-in lengths $s\ge0$, and window sizes $N\ge2$ (Theorems~\ref{thm:4.1}--\ref{thm:4.2}; and up to constants also depending on $c_0,C_0$ in the canonical specialization $c_0N\le s\le C_0N$ of Corollary~\ref{cor:4.5}).

\item \textbf{Stepsize Rules for Spectral Feedback Control:}
Because the tail-averaging filter weights $F_t$ increase during the burn-in phase $1\le t\le s$, standard monotone renewal arguments cannot bound $\chi_{s,N}$. By expanding $F_t$ via a pair-midpoint identity, we establish two concrete constant-stepsize regimes guaranteeing $\chi_{s,N}\le1/2$ (Proposition~\ref{prop:4.3}): (i) for arbitrary spectra $\Hb$ and any finite horizon $T=s+N-1\ge2$, positive correlation of the kernel modes yields $\chi_{s,N}\le\alpha\gamma^2(T-1)\tr(\Hb^2)(1-\gamma\lambda_1)^{-2}$, so any pre-run stepsize $\gamma\le\min\{(2\lambda_1)^{-1},(8\alpha(T-1)\tr(\Hb^2))^{-1/2}\}$ guarantees $\chi_{s,N}\le1/2$; and (ii) splitting the inner convolution at the midpoint bounds $\chi_{s,N}$ through the kernel doubling constant $D_T=\sup_{1\le h\le T}k((h-1)/2)/k(h)$. For polynomially decaying spectra $\theta\Lambda i^{-a}\le\lambda_i\le\Lambda i^{-a}$ ($a>1$), we prove $D_T\le D_*(a,\theta)<\infty$ uniformly in $T$, so a single horizon-independent stepsize ensures $\chi_{s,N}\le1/2$ across all $s,N$, and $\wb_0-\wb^*$ (Corollary~\ref{cor:4.4}).

\item \textbf{Separation Constructions and Conditional Bounds:}
We establish two explicit separation results (Proposition~\ref{prop:4.7}): (i) in two-dimensional noiseless Gaussian regression with $\Hb=\diag(1,(4\gamma N)^{-1})$ and $\wb_0-\wb^*=\eb_1$, burn-in transfer from mode 1 to mode 2 forces excess risk $\Omega(1/(\gamma N^2))$, whereas \citet[Theorem~5.2]{zou2021} gives the exponentially small bound $\cO((1-\gamma)^{2N}/(\gamma^2N^2))$; and (ii) in one-dimensional regression where $k(h)$ decays exponentially and $D_T\to\infty$, two distributions sharing the same $(\Hb,\wb_0-\wb^*,\alpha,\beta)$ exhibit an exponential ratio gap $(r_3/r_2)^N\to\infty$, proving that no single expression depending only on $(\Hb,\wb_0-\wb^*,\gamma,s,N,\alpha,\beta,\sigma_{\noise}^2)$ can match both risks without an additional structural condition (such as $\chi_{s,N}\le\chi_0<1$) or state information (such as $\Bb_s$ in Proposition~\ref{prop:4.8}).
\end{itemize}

\paragraph{Notation.}
We use lowercase letters ($a,b,\gamma,\lambda$) to denote scalars, boldface lowercase letters ($\wb,\xb,\bmeta$) to denote vectors in the separable Hilbert space $\cH$ (where $\cH=\RR^d$ or $\ell_2$), boldface uppercase letters ($\Ab,\Bb,\Hb,\Ib$) to denote bounded self-adjoint linear operators (matrices) on $\cH$, and calligraphic letters ($\cI,\cM,\cS,\cT,\cV$) to denote linear super-operators or scalar risk functionals. For a vector $\wb\in\cH$, we write $(\wb)_i:=\la\vb_i,\wb\ra$ for its $i$-th coordinate in the eigenbasis of $\Hb$, and for a positive semi-definite (PSD) operator $\Ab\succeq\zero$, we write $\norm{\wb}_{\Ab}:=\sqrt{\la\wb,\Ab\wb\ra}$, where $\la\ub,\vb\ra=\ub^\top\vb$. For two vectors $\ub,\vb\in\cH$, their outer product is the rank-one operator $\ub\otimes\vb:=\ub\vb^\top$; for two bounded operators $\Ab,\Bb$ on $\cH$, $\Ab\otimes\Bb$ denotes the linear super-operator acting on operators $\Cb$ via $(\Ab\otimes\Bb)\circ\Cb:=\Ab\Cb\Bb^\top$ (so $(\Hb\otimes\Ib+\Ib\otimes\Hb)\circ\Cb=\Hb\Cb+\Cb\Hb$ preserves self-adjointness); and for two operators $\Ab$ and $\Bb$ whose product $\Ab^\top\Bb$ is trace-class, their trace inner product is $\la\Ab,\Bb\ra:=\tr(\Ab^\top\Bb)$. Given the spectral decomposition $\Hb=\sum_{i\ge1}\lambda_i\vb_i\vb_i^\top$ with $\lambda_1\ge\lambda_2\ge\cdots>0$, for any index pair $0\le k_1\le k_2\le\infty$ and finite head cutoff $0\le k<\infty$, we define the spectral projections and truncated operators
\[
\Ib_{k_1:k_2}:=\sum_{k_1<i\le k_2}\vb_i\vb_i^\top,
\qquad
\Hb_{k_1:k_2}:=\sum_{k_1<i\le k_2}\lambda_i\vb_i\vb_i^\top,
\qquad
\Hb_{0:k}^{-1}:=\sum_{1\le i\le k}\lambda_i^{-1}\vb_i\vb_i^\top,
\]
and write $\norm{\wb}_{\Hb^{-1}}^2:=\sum_{i\ge1}\lambda_i^{-1}(\wb)_i^2\in[0,\infty]$ for the inverse-covariance quadratic form. Throughout the paper, any spectral cutoff $\max\{k:\lambda_k\ge\tau\}$ is set to $0$ when the defining set is empty. For sequences $f_n,g_n\ge0$, we write $f_n\lesssim g_n$ (or $f_n=\cO(g_n)$) if $f_n\le Cg_n$ for a positive constant $C$, $f_n\gtrsim g_n$ (or $f_n=\Omega(g_n)$) if $g_n\lesssim f_n$, and $f_n\asymp g_n$ (or $f_n=\Theta(g_n)$) if both $f_n\lesssim g_n$ and $f_n\gtrsim g_n$ hold. Finally, for real numbers $a,b\in\RR$, we write $a\wedge b:=\min\{a,b\}$ and $a\vee b:=\max\{a,b\}$.

\section{Related Work}\label{sec:2}
\paragraph{Stochastic Approximation and Suffix/Tail Averaging.}
Stochastic gradient descent and Polyak--Ruppert iterate averaging have a rich history in stochastic approximation \citep{robbins1951,polyak1992,ruppert1988,nemirovski2009}. Because full-iterate averaging forgets initial conditions only at a polynomial rate, \citet{rakhlin2012} introduced suffix (tail) averaging for strongly convex stochastic optimization. For least-squares regression in finite and non-parametric dimensions, \citet{bach2013}, \citet{defossez2015}, and \citet{dieuleveut2016,dieuleveut2017} established convergence guarantees for constant- and decaying-stepsize SGD using operator recursions. \citet{jain2017a,jain2017b} developed a Markov chain operator perspective for constant-stepsize least-squares SGD, proving minimax optimal rates and analyzing tail-averaging to achieve exponential forgetting of the initial error in finite dimensions. \citet{jain2018} and \citet{ge2019} further studied accelerated and step-decay schedules for least-squares SGD, while \citet{lakshminarayanan2018} examined constant-stepsize linear stochastic approximation with iterate averaging. In high-dimensional least-squares settings, \citet{paquette2021sgd,paquette2022halting} characterized exact average-case SGD risk trajectories via Volterra integral equations.

\paragraph{Benign Overfitting and Implicit Regularization in Linear Regression.}
The phenomenon of benign overfitting in offline linear regression was formalized by \citet{bartlett2020}, who proved matching upper and lower variance bounds and effective-rank characterizations for the minimum-$\ell_2$-norm interpolator in terms of the data covariance $\Hb$. \citet{tsigler2023} extended these sharp characterizations to ridge regression and minimum-$\ell_2$-norm interpolation, while \citet{hastie2022} and \citet{muthukumar2020} studied asymptotic and finite-sample high-dimensional regression regimes (and \citet{chatterji2021} analyzed overparameterized linear classification). More recently, \citet{wu2025} established instance-wise risk comparisons showing that implicit regularization via gradient methods can dominate explicit ridge regularization in linear regression.

\paragraph{Sharp Risk Bounds for SGD with Weight Averaging in Linear Regression.}
\citet{zou2021} bridged the Markov operator framework of \citet{jain2017a,jain2017b} with the spectral partitioning technique of \citet{bartlett2020} to derive instance-dependent upper and lower bounds for constant-stepsize online SGD with full iterate averaging \citep[Theorems~2.1 and~2.2]{zou2021} and tail-averaging \citep[Theorems~5.1 and~5.2]{zou2021}. Related works have studied several complementary directions: \citet{berthier2020} studied tight non-parametric rates in noiseless linear models; \citet{wu2022} analyzed the last-iterate risk of SGD with polynomial and step-decay stepsize schedules; \citet{li2023} established instance-dependent risk bounds for accelerated SGD with Nesterov momentum and tail averaging; \citet{lin2024} and \citet{li2025functional} studied scaling laws and learning-rate schedules under power-law spectral conditions; and \citet{li2025} analyzed SGD with exponential moving average (EMA), where a parallel gap between upper and lower bounds on the head initialization subspace appears between their Theorems~4.1 and~4.2. In large-scale deep learning and language model pretraining, checkpoint-based tail averaging---termed Latest Weight Averaging (LAWA) \citep{kaddour2022,sanyal2023}---has been used widely to accelerate convergence, motivating a sharp theoretical understanding of tail-averaged SGD. Methodologically, \citet{meterez2025} recently simplified the proof of \citet{zou2021} by working in the diagonal eigenbasis of $\Hb$ while retaining the same gap on the head initialization subspace, and \citet{meterez2026} studied horizon-free learning-rate schedules with weight averaging for anytime pretraining.

\section{Problem Setup and Preliminaries}\label{sec:3}
We consider linear regression over a separable Hilbert space $\cH$ (which may be $\RR^d$ or infinite-dimensional). Let $\cD$ be an unknown population distribution over feature-response pairs $(\xb,y)\in\cH\times\RR$. The population risk of a weight vector $\wb\in\cH$ is given by the expected squared loss
\[
L(\wb)=\frac12\EE_{(\xb,y)\sim\cD}\bigl[\bigl(y-\la\wb,\xb\ra\bigr)^2\bigr].
\]

\subsection{Tail-Averaged Stochastic Gradient Descent}\label{sec:3.1}
Given an initialization $\wb_0\in\cH$ and a stream of independent and identically distributed (i.i.d.) samples $\{(\xb_t,y_t)\}_{t\ge1}$ drawn from $\cD$, constant-stepsize SGD updates the weight vector at iteration $t\ge1$ according to
\begin{equation}\label{eq:3.1}
\wb_t=\wb_{t-1}+\gamma\bigl(y_t-\la\wb_{t-1},\xb_t\ra\bigr)\xb_t,
\end{equation}
where $\gamma>0$ is the learning rate (stepsize). Given a burn-in period of length $s\ge0$ and an averaging window of length $N\ge1$, SGD with tail-averaging outputs the average of the iterates from step $s$ to step $s+N-1$:
\begin{equation}\label{eq:3.2}
\overbar{\wb}_{s:s+N}=\frac1N\sum_{t=s}^{s+N-1}\wb_t.
\end{equation}
During the first $s$ burn-in iterations, the algorithm executes constant-stepsize SGD updates \eqref{eq:3.1} without recording the iterates into the running average; from iteration $t=s$ to $s+N-1$, it averages the subsequent $N$ iterates to form $\overbar{\wb}_{s:s+N}$ in \eqref{eq:3.2}. When $s=0$, \eqref{eq:3.2} reduces to standard full-iterate averaging $\overbar{\wb}_{0:N}=\frac1N\sum_{t=0}^{N-1}\wb_t$ \citep{zou2021}.

\subsection{Model Assumptions}\label{sec:3.2}
Following \citet{zou2021}, we adopt standard second- and fourth-moment regularity conditions on the feature distribution and label noise. Specifically, our upper bounds rely solely on Assumptions~\ref{ass:3.1}, \ref{ass:3.2}, and~\ref{ass:3.3}, while our lower bounds rely on the well-specified linear model \eqref{eq:3.4} together with Assumptions~\ref{ass:3.1}, \ref{ass:3.2}, and~\ref{ass:3.4}.

\begin{assumption}[Regularity Conditions; \citealt{zou2021}, Assumption~2.1]\label{ass:3.1}
The second moment operator $\Hb:=\EE_{(\xb,y)\sim\cD}[\xb\xb^\top]$ is strictly positive definite ($\Hb\succ\zero$) and has finite trace $\tr(\Hb)<\infty$. Moreover, the population risk $L(\wb)$ admits a unique global minimizer $\wb^*=\argmin_{\wb\in\cH}L(\wb)$ satisfying $L(\wb^*)<\infty$.
\end{assumption}

Let $\Hb=\sum_{i\ge1}\lambda_i\vb_i\vb_i^\top$ be the eigendecomposition of $\Hb$, where the eigenvalues are ordered non-increasingly as $\lambda_1\ge\lambda_2\ge\cdots>0$ and $\sum_{i\ge1}\lambda_i=\tr(\Hb)<\infty$. To control the fourth-moment behavior of the stochastic rank-one updates $\xb_t\xb_t^\top$, define the fourth-moment PSD super-operators $\cM$ and $\wt{\cM}$ acting on any symmetric operator $\Ab$ by
\begin{equation}\label{eq:3.3}
\cM\circ\Ab:=\EE_{(\xb,y)\sim\cD}\bigl[\xb\xb^\top\Ab\xb\xb^\top\bigr],
\qquad\wt{\cM}\circ\Ab:=\Hb\Ab\Hb.
\end{equation}

\begin{assumption}[Fourth-Moment Upper Bound; \citealt{zou2021}, Assumption~2.2]\label{ass:3.2}
There exists a constant $\alpha>0$ such that for any PSD operator $\Ab\succeq\zero$, it holds that
\[
\cM\circ\Ab=\EE_{(\xb,y)\sim\cD}\bigl[\xb\xb^\top\Ab\xb\xb^\top\bigr]\preceq\alpha\tr(\Hb\Ab)\Hb.
\]
\end{assumption}

As shown in Lemma~\ref{lem:C.1} (restated from \citet[Lemma~A.1]{zou2021}), Assumption~\ref{ass:3.2} holds with $\alpha=16\sigma_z^4$ whenever the standardized projections of $\xb$ are $\sigma_z^2$-sub-Gaussian (i.e., $\EE[\exp(\la\ub,\xb\ra)]\le\exp(\sigma_z^2\norm{\ub}_{\Hb}^2/2)$ for all $\ub\in\cH$, or $\zb:=\Hb^{-1/2}\xb$ is $\sigma_z^2$-sub-Gaussian in finite dimension), and with $\alpha=3$ when $\xb\sim\cN(\zero,\Hb)$. Note also that taking $\Ab=\vb_1\vb_1^\top$ in $\Hb\Ab\Hb=\wt{\cM}\circ\Ab\preceq\cM\circ\Ab\preceq\alpha\tr(\Hb\Ab)\Hb$ yields $\lambda_1^2\le\alpha\lambda_1^2$, so $\alpha\ge1$ always holds; moreover, $(\cM-\wt{\cM})\circ\Ab\preceq\cM\circ\Ab\preceq\alpha\tr(\Hb\Ab)\Hb$ for all $\Ab\succeq\zero$.

\begin{assumption}[Noise Condition; \citealt{zou2021}, Assumption~2.3]\label{ass:3.3}
Define the residual noise $\xi:=y-\la\wb^*,\xb\ra$ and the noise-weighted covariance operator $\bSigma:=\EE_{(\xb,y)\sim\cD}[\xi^2\xb\xb^\top]$. Suppose that
\[
\sigma^2:=\nbr{\Hb^{-1/2}\bSigma\Hb^{-1/2}}_2<\infty,
\]
or equivalently, $\bSigma\preceq\sigma^2\Hb$ in the Loewner PSD order.
\end{assumption}

Assumption~\ref{ass:3.3} accommodates model misspecification and heteroscedastic noise, requiring only that $\EE[\xi\xb]=\nabla L(\wb^*)=\zero$ and $\bSigma\preceq\sigma^2\Hb$. For the lower bound analysis, we consider the well-specified homoscedastic linear regression model (slightly relaxing \citet[Eq.~(2.1)]{zou2021}, which specifies Gaussian noise $\xi\sim\cN(0,\sigma_{\noise}^2)$, to arbitrary centered finite-variance noise independent of $\xb$):
\begin{equation}\label{eq:3.4}
y=\la\wb^*,\xb\ra+\xi,
\qquad\text{where }\xi\perp\xb,\quad
\EE[\xi]=0,\quad\text{and}\quad\EE[\xi^2]=\sigma_{\noise}^2<\infty.
\end{equation}
Under \eqref{eq:3.4}, $\bSigma=\sigma_{\noise}^2\Hb$, so Assumption~\ref{ass:3.3} holds with equality ($\sigma^2=\sigma_{\noise}^2$). Finally, to establish lower bounds on the fourth-moment bias-variance interaction, we adopt the fourth-moment lower bound condition of \citet{zou2021}.

\begin{assumption}[Fourth-Moment Lower Bound; \citealt{zou2021}, Assumption~2.4]\label{ass:3.4}
There exists a constant $\beta\in(0,\alpha]$ such that for any PSD operator $\Ab\succeq\zero$, it holds that
\[
(\cM-\wt{\cM})\circ\Ab=\EE_{(\xb,y)\sim\cD}\bigl[\xb\xb^\top\Ab\xb\xb^\top\bigr]-\Hb\Ab\Hb\succeq\beta\tr(\Hb\Ab)\Hb.
\]
\end{assumption}

When $\xb\sim\cN(\zero,\Hb)$, Isserlis' theorem yields $\cM\circ\Ab=2\Hb\Ab\Hb+\tr(\Hb\Ab)\Hb$, so $(\cM-\wt{\cM})\circ\Ab=\Hb\Ab\Hb+\tr(\Hb\Ab)\Hb\succeq\tr(\Hb\Ab)\Hb$ and Assumption~\ref{ass:3.4} holds with $\beta=1$ (and with $\beta=2$ in scalar Gaussian regression $d=1$).

\subsection{Bias-Variance Decomposition and Effective Risk Components}\label{sec:3.3}
Following \citet{jain2017b}, \citet{zou2021}, and \citet{li2025}, let $\bmeta_t:=\wb_t-\wb^*$ denote the estimation error at iteration $t$ and $\overbar{\bmeta}_{s:s+N}:=\overbar{\wb}_{s:s+N}-\wb^*=\frac1N\sum_{t=s}^{s+N-1}\bmeta_t$. By linearity of \eqref{eq:3.1}, $\bmeta_t=\bmeta_t^{\bias}+\bmeta_t^{\variance}$, where the noiseless \emph{bias process} and zero-initialized \emph{variance process} satisfy
\begin{align}
\bmeta_t^{\bias}&=\bigl(\Ib-\gamma\xb_t\xb_t^\top\bigr)\bmeta_{t-1}^{\bias},
&\bmeta_0^{\bias}&=\wb_0-\wb^*,\label{eq:5.1}\\
\bmeta_t^{\variance}&=\bigl(\Ib-\gamma\xb_t\xb_t^\top\bigr)\bmeta_{t-1}^{\variance}+\gamma\xi_t\xb_t,
&\bmeta_0^{\variance}&=\zero.\label{eq:5.2}
\end{align}
Here $\xi_t:=y_t-\la\wb^*,\xb_t\ra$, and the single-step second-moment operators are $\Bb_t:=\EE[\bmeta_t^{\bias}\otimes\bmeta_t^{\bias}]$ and $\Cb_t:=\EE[\bmeta_t^{\variance}\otimes\bmeta_t^{\variance}]$ (initialized at $\Bb_0=(\wb_0-\wb^*)\otimes(\wb_0-\wb^*)$ and $\Cb_0=\zero$). Letting $\overbar{\bmeta}_{s:s+N}^{\bias}:=\frac1N\sum_{t=s}^{s+N-1}\bmeta_t^{\bias}$ and $\overbar{\bmeta}_{s:s+N}^{\variance}:=\frac1N\sum_{t=s}^{s+N-1}\bmeta_t^{\variance}$, Lemma~\ref{lem:C.3} \citep{jain2017b,zou2021} shows that $\EE[L(\overbar{\wb}_{s:s+N})]-L(\wb^*)\le2\cdot\bias+2\cdot\variance$ under Assumptions~\ref{ass:3.1}--\ref{ass:3.3}, and $\EE[L(\overbar{\wb}_{s:s+N})]-L(\wb^*)=\bias+\variance$ under the well-specified model \eqref{eq:3.4}, where
\begin{equation}\label{eq:5.6}
\bias:=\frac12\inner{\Hb}{\EE\bigl[\overbar{\bmeta}_{s:s+N}^{\bias}\otimes\overbar{\bmeta}_{s:s+N}^{\bias}\bigr]},\qquad
\variance:=\frac12\inner{\Hb}{\EE\bigl[\overbar{\bmeta}_{s:s+N}^{\variance}\otimes\overbar{\bmeta}_{s:s+N}^{\variance}\bigr]}.
\end{equation}
Because the rank-one updates $\xb_t\xb_t^\top$ in \eqref{eq:5.1} fluctuate around $\Hb=\EE[\xb_t\xb_t^\top]$, the bias process has mean $\EE[\bmeta_t^{\bias}]=(\Ib-\gamma\Hb)^t(\wb_0-\wb^*)$ together with stochastic fluctuations $\bmeta_t^{\bias}-\EE[\bmeta_t^{\bias}]$ driven by multiplicative gradient noise $(\xb_t\xb_t^\top-\Hb)\bmeta_{t-1}^{\bias}$. Following \citet{zou2021} and \citet{li2025}, we therefore decompose $\bias+\variance=\EffectiveBias+\EffectiveVar$ into the deterministic drift bias $\EffectiveBias$ and the total stochastic variance error $\EffectiveVar$:
\begin{align}
\EffectiveBias&:=\frac12\norm{\EE\bigl[\overbar{\bmeta}_{s:s+N}^{\bias}\bigr]}_{\Hb}^2
=\sum_{i=1}^{\infty}\frac{(\wb_0-\wb^*)_i^2}{2\gamma^2N^2\lambda_i}(1-\gamma\lambda_i)^{2s}\bigl(1-(1-\gamma\lambda_i)^N\bigr)^2,\label{eq:4.1}\\
\EffectiveVar&:=(\bias-\EffectiveBias)+\variance
=\frac12\EE\norm{\overbar{\bmeta}_{s:s+N}^{\bias}-\EE\bigl[\overbar{\bmeta}_{s:s+N}^{\bias}\bigr]}_{\Hb}^2+\variance,\label{eq:effvar_def}
\end{align}
where $\EffectiveBias=\frac12\norm{\frac1N\sum_{t=s}^{s+N-1}(\Ib-\gamma\Hb)^t(\wb_0-\wb^*)}_{\Hb}^2$ is the excess risk of deterministic gradient descent averaged over $[s,s+N-1]$. The effective variance in~\eqref{eq:effvar_def} is the sum of the bias-induced variance $\bias-\EffectiveBias$ and the additive-noise variance $\variance$.

\section{Main Results}\label{sec:4}
In this section, we present our main upper and lower bounds on the excess risk $\EE[L(\overbar{\wb}_{s:s+N})]-L(\wb^*)$ of SGD with tail-averaging (Section~\ref{sec:4.1}), establish concrete stepsize rules guaranteeing the spectral feedback condition for general and polynomially decaying spectra (Section~\ref{sec:4.2}), specialize our bounds to the canonical regime $s=\Theta(N)$ to compare directly with \citet{zou2021} (Section~\ref{sec:4.3}), and present our separation constructions and conditional bounds (Section~\ref{sec:4.4}).

Before stating our theorems, recall from Section~\ref{sec:3.3} the deterministic drift bias $\EffectiveBias$ in \eqref{eq:4.1} and the effective variance $\EffectiveVar$ in \eqref{eq:effvar_def}, and let $T:=s+N-1$ denote the final iteration index in \eqref{eq:3.2}. For each step $t\in\{1,\ldots,T\}$ and real lag $h\ge0$, we define the tail-averaging filter polynomial $q_t(\lambda)$, the receiving sensitivity weight $F_t$, and the propagation kernel $k(h)$ by\footnote{Throughout Section~\ref{sec:4}, an empty sum for $T=0$ or an empty maximum over $1\le r<T$ for $T\le1$ is $0$.}
\begin{equation}\label{eq:4.2}
q_t(\lambda):=\frac1N\sum_{m=s\vee t}^{T}(1-\gamma\lambda)^{m-t},\qquad
F_t:=\sum_{i=1}^{\infty}\lambda_i^2q_t(\lambda_i)^2,\qquad
k(h):=\sum_{i=1}^{\infty}\lambda_i^2(1-\gamma\lambda_i)^{2h},
\end{equation}
together with the \emph{spectral feedback coefficient}
\begin{equation}\label{eq:4.3}
\chi_{s,N}:=\alpha\gamma^2\max_{1\le r<T}\frac{\sum_{t=r+1}^{T}k(t-r-1)F_t}{F_r}.
\end{equation}
Here $F_t$ measures the sensitivity of $\overbar{\wb}_{s:s+N}$ to a stochastic gradient perturbation at step $t$, while $\chi_{s,N}$ measures the worst-case single-hop amplification of fourth-moment energy across the trajectory (see Section~\ref{sec:5.1} for a detailed derivation).

\subsection{Upper and Lower Bounds of Excess Risk}\label{sec:4.1}
We first present our upper bound on the excess risk of SGD with tail-averaging.

\begin{theorem}[Upper Bound]\label{thm:4.1}
Consider SGD with tail-averaging \eqref{eq:3.2}. Suppose Assumptions~\ref{ass:3.1}, \ref{ass:3.2}, and~\ref{ass:3.3} hold, and that the stepsize and spectral feedback coefficient satisfy $\gamma\le1/(2\lambda_1)$ and $\chi_{s,N}<1$. Then for any $s\ge0$ and $N\ge1$, the excess risk can be upper bounded as follows:
\[
\EE\bigl[L(\overbar{\wb}_{s:s+N})\bigr]-L(\wb^*)\le2\cdot\EffectiveBias+2\cdot\EffectiveVar,
\]
where the factor of $2$ can be omitted under the well-specified model \eqref{eq:3.4}, the effective bias in \eqref{eq:4.1} satisfies
\begin{equation}\label{eq:4.4}
\EffectiveBias\le\frac{1}{2\gamma^2N^2}\cdot\norm{(\Ib-\gamma\Hb)^s(\wb_0-\wb^*)}_{\Hb^{-1}_{0:k^*}}^2
+\frac12\cdot\norm{(\Ib-\gamma\Hb)^s(\wb_0-\wb^*)}_{\Hb_{k^*:\infty}}^2,
\end{equation}
and the effective variance in \eqref{eq:effvar_def} satisfies\footnote{In \eqref{eq:4.5} and \eqref{eq:4.8}, we adopt the standard conventions $1/(\gamma|\lambda_i-\lambda_j|):=+\infty$ when $\lambda_i=\lambda_j$ and $s^{-1}:=+\infty$ when $s=0$ (so the burn-in double sum over $i,j$ vanishes at $s=0$).}
\begingroup\medmuskip=3mu\thickmuskip=4mu
\begin{align}
\EffectiveVar\le{}&\frac{\alpha\bigl(\norm{(\Ib-\gamma\Hb)^s(\wb_0-\wb^*)}_{\Ib_{0:k^*}}^2+N\gamma\norm{(\Ib-\gamma\Hb)^s(\wb_0-\wb^*)}_{\Hb_{k^*:\infty}}^2\bigr)}{N\gamma(1-\chi_{s,N})}\Bigl(\frac{k^*}{N}+N\gamma^2\sum_{i>k^*}\lambda_i^2\Bigr)\notag\\
&+\frac{\alpha}{1-\chi_{s,N}}\sum_{i,j=1}^{\infty}\min\Bigl\{\frac1{N^2},\gamma^2\lambda_i^2\Bigr\}\frac{\lambda_j(\wb_0-\wb^*)_j^2(1-\gamma(\lambda_i\wedge\lambda_j))^{2(s-1)}}{s^{-1}\vee\gamma|\lambda_i-\lambda_j|}\notag\\
&+\frac{\sigma^2}{1-\chi_{s,N}}\cdot\left(\frac{k^*}{N}+\gamma\sum_{k^*<i\le k^\dagger}\lambda_i+(s+N)\gamma^2\sum_{i>k^\dagger}\lambda_i^2\right),\label{eq:4.5}
\end{align}
\endgroup
where the eigenvalue cutoffs are defined as
\begin{equation}\label{eq:4.6}
k^*=\max\Bigl\{k:\lambda_k\ge\frac1{\gamma N}\Bigr\},\qquad
k^\dagger=\max\Bigl\{k:\lambda_k\ge\frac1{\gamma(s+N)}\Bigr\}.
\end{equation}
\end{theorem}

We next establish our matching lower bound on the excess risk of SGD with tail-averaging:

\begin{theorem}[Lower Bound]\label{thm:4.2}
Consider SGD with tail-averaging \eqref{eq:3.2} under the well-specified model \eqref{eq:3.4}. Suppose Assumptions~\ref{ass:3.1}, \ref{ass:3.2}, and~\ref{ass:3.4} hold, and the hyperparameters satisfy $\gamma\le1/(2\lambda_1)$ and $N\ge2$. Then for any $s\ge0$, the excess risk decomposes as
\[
\EE\bigl[L(\overbar{\wb}_{s:s+N})\bigr]-L(\wb^*)=\EffectiveBias+\EffectiveVar,
\]
where the effective bias in \eqref{eq:4.1} satisfies
\begin{equation}\label{eq:4.7}
\EffectiveBias\ge\frac1{8\gamma^2N^2}\cdot\norm{(\Ib-\gamma\Hb)^s(\wb_0-\wb^*)}_{\Hb^{-1}_{0:k^*}}^2
+\frac18\cdot\norm{(\Ib-\gamma\Hb)^s(\wb_0-\wb^*)}_{\Hb_{k^*:\infty}}^2,
\end{equation}
and the effective variance in \eqref{eq:effvar_def} satisfies
\begingroup\medmuskip=3mu\thickmuskip=4mu
\begin{align}
\EffectiveVar\ge{}&\frac{\beta\bigl(\norm{(\Ib-\gamma\Hb)^s(\wb_0-\wb^*)}_{\Ib_{0:k^*}}^2+N\gamma\norm{(\Ib-\gamma\Hb)^s(\wb_0-\wb^*)}_{\Hb_{k^*:\infty}}^2\bigr)}{864N\gamma}\Bigl(\frac{k^*}{N}+N\gamma^2\sum_{i>k^*}\lambda_i^2\Bigr)\notag\\
&+\frac\beta{108}\sum_{i,j=1}^{\infty}\min\Bigl\{\frac1{N^2},\gamma^2\lambda_i^2\Bigr\}\frac{\lambda_j(\wb_0-\wb^*)_j^2(1-\gamma(\lambda_i\wedge\lambda_j))^{2(s-1)}}{s^{-1}\vee\gamma|\lambda_i-\lambda_j|}\notag\\
&+\frac{\sigma_{\noise}^2}{216}\cdot\left(\frac{k^*}{N}+\gamma\sum_{k^*<i\le k^\dagger}\lambda_i+(s+N)\gamma^2\sum_{i>k^\dagger}\lambda_i^2\right),\label{eq:4.8}
\end{align}
\endgroup
with $k^*$ and $k^\dagger$ defined in \eqref{eq:4.6}.
\end{theorem}

\paragraph{Structure and Interpretation of the Bounds.}
The proofs of Theorems~\ref{thm:4.1} and~\ref{thm:4.2} are given in Appendix~\ref{app:B}. The bounds on $\EffectiveBias$ and $\EffectiveVar$ separate the following contributions:
\begin{itemize}
\item \emph{Effective Bias:} In \eqref{eq:4.4} and \eqref{eq:4.7}, the initial error $\wb_0-\wb^*$ is first contracted by $(\Ib-\gamma\Hb)^s$ during the $s$ burn-in steps, and then enjoys $\cO((\gamma^2N^2\lambda_i)^{-1})$ averaging decay on the head subspace $\Ib_{0:k^*}$ while remaining unaveraged on the tail subspace $\Hb_{k^*:\infty}$.
\item \emph{Effective Variance:} Both \eqref{eq:4.5} and \eqref{eq:4.8} consist of three matching components: (i) the \emph{averaging-window multiplicative variance} (first line), which has the exact form of full-averaging SGD restarted from the burn-in drifted vector $(\Ib-\gamma\Hb)^s(\wb_0-\wb^*)$ at step $s$; (ii) the \emph{burn-in cross-spectral transfer variance} (second line), where initial error in source mode $j$ scatters fourth-moment energy into mode $i$ during the $s$ burn-in steps with effective interaction duration $\min\{s,(\gamma|\lambda_i-\lambda_j|)^{-1}\}$ and slower-mode damping $(1-\gamma(\lambda_i\wedge\lambda_j))^{2(s-1)}$; and (iii) the \emph{additive label-noise variance} (third line), partitioned across $i\le k^*$, $k^*<i\le k^\dagger$, and $i>k^\dagger$.
\end{itemize}
Whenever $\chi_{s,N}\le\chi_0<1$ for a constant $\chi_0<1$, every term in Theorems~\ref{thm:4.1} and~\ref{thm:4.2} matches up to a multiplicative factor depending only on $\alpha$, $\beta$, and $1-\chi_0$ across all dimensions, spectra, initializations, and horizons $s\ge0$, $N\ge2$. Moreover, when $s=0$, the burn-in sum vanishes and Theorems~\ref{thm:4.1} and~\ref{thm:4.2} recover \citet[Theorems~2.1 and~2.2]{zou2021} (while extending the lower bound from $N\ge500$ to all $N\ge2$).

\subsection{Guaranteeing the Spectral Feedback Condition \texorpdfstring{$\chi_{s,N}\le\chi_0<1$}{chi(s,N) <= chi0 < 1}}\label{sec:4.2}
We now show how the spectral feedback condition $\chi_{s,N}\le1/2$ can be guaranteed prior to execution, both for arbitrary covariance spectra $\Hb$ (under a horizon-dependent constant stepsize) and for polynomially decaying spectra $\lambda_i\asymp i^{-a}$ (under a single horizon-independent constant stepsize). Recall $k(h)=\sum_{i\ge1}\lambda_i^2(1-\gamma\lambda_i)^{2h}$ for real $h\ge0$, and define its doubling constant over $[1,T]$ by
\begin{equation}\label{eq:4.9}
D_T:=\sup_{1\le h\le T}\frac{k((h-1)/2)}{k(h)}.
\end{equation}

\begin{proposition}[Pair-Midpoint Convolution Bounds on $\chi_{s,N}$ and General-Spectrum Stepsize Rule]\label{prop:4.3}
Suppose Assumption~\ref{ass:3.1} holds and $0<\gamma<1/\lambda_1$. Then for every $s\ge0$ and $N\ge1$ (with $T=s+N-1\ge0$), the receiving weights $F_t$ in \eqref{eq:4.2} satisfy the exact pair-midpoint identity
\begin{equation}\label{eq:4.10}
F_t=\frac1{N^2}\sum_{a=s\vee t}^{T}\sum_{b=s\vee t}^{T}k\Bigl(\frac{a+b}{2}-t\Bigr),\qquad\forall t\in\{1,\ldots,T\}.
\end{equation}
Consequently, $\chi_{s,N}=0$ whenever $T\le1$, and for all $T=s+N-1\ge2$, the spectral feedback coefficient $\chi_{s,N}$ in \eqref{eq:4.3} satisfies both:
\begin{enumerate}
\item \textup{(General-Spectrum Horizon-Dependent Bound):} For any covariance operator $\Hb$ and $T\ge2$,
\begin{equation}\label{eq:4.11}
\chi_{s,N}\le\frac{\alpha\gamma^2(T-1)\tr(\Hb^2)}{(1-\gamma\lambda_1)^2},
\end{equation}
so choosing any constant pre-run stepsize satisfying
\begin{equation}\label{eq:4.12}
\gamma\le\min\left\{\frac1{2\lambda_1},\frac1{\sqrt{8\alpha(T-1)\tr(\Hb^2)}}\right\}\quad\Longrightarrow\quad\chi_{s,N}\le\frac12.
\end{equation}
\item \textup{(Kernel-Doubling Bound):} In terms of the doubling constant $D_T$ in \eqref{eq:4.9},
\begin{equation}\label{eq:4.13}
\chi_{s,N}\le2\alpha D_T\gamma^2\sum_{n=0}^{\infty}k(n)=2\alpha D_T\gamma\sum_{i=1}^{\infty}\frac{\lambda_i}{2-\gamma\lambda_i}\le\frac{2\alpha D_T\gamma\tr(\Hb)}{2-\gamma\lambda_1}.
\end{equation}
\end{enumerate}
\end{proposition}

Proposition~\ref{prop:4.3}(1) guarantees $\chi_{s,N}\le1/2$ (and hence matching upper and lower bounds in Theorems~\ref{thm:4.1} and~\ref{thm:4.2}) for \emph{arbitrary} spectra $\Hb$ under the horizon-dependent constant stepsize \eqref{eq:4.12}. Moreover, when the eigenvalues of $\Hb$ decay polynomially as $\lambda_i\asymp i^{-a}$ ($a>1$)---the standard power-law family studied in \citet{caponnetto2007,dieuleveut2016,bartlett2020,zou2021,li2025}---the propagation kernel decays polynomially as $k(h)\asymp\Lambda^2(1+\gamma\Lambda h)^{-(2-1/a)}$. In this regime, Proposition~\ref{prop:4.3}(2) strengthens \eqref{eq:4.12} to a \emph{single horizon-independent} stepsize $\gamma$ ensuring $\chi_{s,N}\le1/2$ across all $s$ and $N$.

\begin{corollary}[Uniform Horizon-Free Matching under Polynomial Eigenvalue Decay]\label{cor:4.4}
Suppose there exist constants $a>1$, $\Lambda>0$, and $\theta\in(0,1]$ such that $\theta\Lambda i^{-a}\le\lambda_i\le\Lambda i^{-a}$ for all $i\ge1$ (in infinite dimension) or for $1\le i\le d$ (in finite dimension $d\ge2$, over horizons $\gamma\Lambda T\le(d/2)^a$). Then there exists a finite constant $D_*=D_*(a,\theta)\ge1$ such that $\gamma\Lambda\le1/2$ implies $D_T\le D_*$, and hence any fixed stepsize satisfying
\begin{equation}\label{eq:4.14}
\gamma\le\min\left\{\frac1{2\Lambda},\frac1{4\alpha D_*\tr(\Hb)}\right\}
\end{equation}
ensures $\chi_{s,N}\le1/2$. Consequently, under \eqref{eq:3.4} and Assumptions~\ref{ass:3.1}--\ref{ass:3.4}, the upper bound of Theorem~\ref{thm:4.1} and the lower bound of Theorem~\ref{thm:4.2} match up to constants depending only on $\alpha$ and $\beta$ uniformly over all $\wb_0-\wb^*$, $s\ge0$, and $N\ge2$ (and Corollary~\ref{cor:4.5} below holds whenever $c_0N\le s\le C_0N$ up to constants depending only on $\alpha,\beta,c_0,C_0$).
\end{corollary}

\subsection{Canonical Regime \texorpdfstring{$s=\Theta(N)$}{s = Theta(N)} and Comparison with \texorpdfstring{\citet{zou2021}}{Zou et al. (2021)}}\label{sec:4.3}
In the standard tail-averaging regime where $s=\Theta(N)$ (for instance, $s=N$, which averages the second half of a $2N$-step run as in \citet{jain2018} and \citet[Section~5]{zou2021}), the eigenvalue cutoffs $k^*$ and $k^\dagger$ have thresholds $(\gamma N)^{-1}$ and $(\gamma(s+N))^{-1}$ within a constant factor of each other, so $\norm{\wb_0-\wb^*}_{\Hb_{k^*:\infty}}^2\asymp\norm{(\Ib-\gamma\Hb)^s(\wb_0-\wb^*)}_{\Hb_{k^*:\infty}}^2$ and $\gamma\lambda_i\asymp N\gamma^2\lambda_i^2\asymp N^{-1}$ for all $k^*<i\le k^\dagger$. Splitting the burn-in variance sum in \eqref{eq:4.5} and \eqref{eq:4.8} into fast index modes ($i\le k^*$, where $\min\{N^{-2},\gamma^2\lambda_i^2\}=N^{-2}$) and slow index modes ($i>k^*$, where $\min\{N^{-2},\gamma^2\lambda_i^2\}=\gamma^2\lambda_i^2$) yields the following corollary and the term-by-term comparison in Table~\ref{tab:1}.

\begin{corollary}[Matching Excess Risk Bounds when $s=\Theta(N)$]\label{cor:4.5}
Suppose $c_0N\le s\le C_0N$ for constants $0<c_0\le C_0<\infty$, and suppose the assumptions of Theorems~\ref{thm:4.1} and~\ref{thm:4.2} hold with $\chi_{s,N}\le\chi_0<1$. Then the excess risk satisfies
\begin{equation}\label{eq:4.15}
\EffectiveBias=\Theta\left(\frac1{\gamma^2N^2}\cdot\norm{(\Ib-\gamma\Hb)^s(\wb_0-\wb^*)}_{\Hb^{-1}_{0:k^*}}^2+\norm{\wb_0-\wb^*}_{\Hb_{k^*:\infty}}^2\right),
\end{equation}
\begin{align}
\EffectiveVar=\Theta\Biggl(&\thupurple{\frac{\norm{\wb_0-\wb^*}_{\Ib_{0:k^*}}^2+N\gamma\norm{\wb_0-\wb^*}_{\Hb_{k^*:\infty}}^2}{N\gamma}\cdot\left(N\gamma^2\sum_{i>k^*}\lambda_i^2\right)}\notag\\*
&+\frac{\norm{(\Ib-\gamma\Hb)^s(\wb_0-\wb^*)}_{\Ib_{0:k^*}}^2+N\gamma\norm{\wb_0-\wb^*}_{\Hb_{k^*:\infty}}^2}{N\gamma}\cdot\frac{k^*}{N}\notag\\*
&+\frac1{N^2}\sum_{i\le k^*}\sum_{j\le k^*}\lambda_j(\wb_0-\wb^*)_j^2\bigl(1-\gamma(\lambda_i\wedge\lambda_j)\bigr)^{2(s-1)}\min\left\{s,\frac1{\gamma|\lambda_i-\lambda_j|}\right\}\notag\\*
&+\sigma_{\noise}^2\cdot\left(\frac{k^*}{N}+N\gamma^2\sum_{i>k^*}\lambda_i^2\right)\Biggr),\label{eq:4.16}
\end{align}
where the constants hidden in $\Theta(\cdot)$ depend only on $\alpha$, $\beta$, $1-\chi_0$, $c_0$, and $C_0$.
\end{corollary}

\begin{remark}[Resolution of the Total-Risk Gap in \citet{zou2021} and Role of Multi-Hop Return Paths]\label{rem:4.6}
Comparing \eqref{eq:4.16} with \citet[Theorems~5.1 and~5.2]{zou2021} (see Table~\ref{tab:1}) clarifies both why the rank-one factored form in prior work could not match and how the terms of \eqref{eq:4.16} combine in the total excess risk:
\begin{enumerate}
\item \emph{Presence of Slow Spectral Modes ($k^*<d$, so $\sum_{i>k^*}\lambda_i^2>0$):} In Panel~B ($i>k^*$), the first line of \eqref{eq:4.16} retains the undamped head initialization contribution $\frac1{\gamma N}\norm{\wb_0-\wb^*}_{\Ib_{0:k^*}}^2\cdot\bigl(N\gamma^2\sum_{i>k^*}\lambda_i^2\bigr)$, which is missing from \citet[Theorem~5.2]{zou2021}. In multi-dimensional problems with $k^*<d$, energy scattered into slow modes $i>k^*$ can also scatter back into fast modes $i\le k^*$; under $\chi_{s,N}\le\chi_0<1$, this \thupurple{multi-hop return contribution} is absorbed into the total-risk bound via $(1-\chi_{s,N})^{-1}$ multiplying the full single-hop expression in \eqref{eq:4.16}.\footnote{Accordingly, Panels~A and~B in Table~\ref{tab:1} record the fast-mode ($i\le k^*$) and slow-mode ($i>k^*$) index components of the single-hop functional $\cS_{s,N}$ in the total excess risk.}
\item \emph{Well-Resolved Spectral Regime ($d\le k^*$, so $\Hb_{k^*:\infty}=\zero$):} Conversely, when all eigenvalues satisfy $\lambda_i\ge(\gamma N)^{-1}$, the slow subspace is empty ($\sum_{i>k^*}\lambda_i^2=0$), so Panel~B vanishes and no slow-mode return path exists. Every remaining bias-induced variance term in Panel~A (the second and third lines of \eqref{eq:4.16}) is exponentially damped in $s$ by $(1-\gamma(\lambda_i\wedge\lambda_j))^{2(s-1)}$, eliminating the undamped $\frac1{\gamma N}\norm{\wb_0-\wb^*}_{\Ib_{0:k^\dagger}}^2\cdot\frac{k^*}{N}$ artifact in \citet[Theorem~5.1]{zou2021}.
\end{enumerate}
\end{remark}

\begin{table}[t]
\centering
\caption{Comparison of the fast-mode ($i\le k^*$) and slow-mode ($i>k^*$) index components in the total bias-induced variance bound for tail-averaged SGD $\overbar{\wb}_{s:s+N}$ in the canonical regime $c_0N\le s\le C_0N$, omitting multiplicative constants depending on $\alpha,\beta,c_0,C_0$, and the stability margin ($1-\gamma\alpha\tr(\Hb)$ in \citet[Theorem~5.1]{zou2021}, and $1-\chi_0$ in Corollary~\ref{cor:4.5} under $\chi_{s,N}\le\chi_0<1$).}\label{tab:1}
\small
\setlength{\tabcolsep}{4pt}
\renewcommand{\arraystretch}{1.5}
\begin{tabular}{@{}l l@{}}
\toprule
\textbf{Bound}&\textbf{Panel A: Fast-Mode Index Component ($i\le k^*$)}\\
\midrule
\citet[Theorem~5.1]{zou2021} (Upper)&$\displaystyle\frac{\norm{\wb_0-\wb^*}_{\Ib_{0:k^\dagger}}^2+N\gamma\norm{\wb_0-\wb^*}_{\Hb_{k^\dagger:\infty}}^2}{N\gamma}\cdot\frac{k^*}{N}$\\[6pt]
\citet[Theorem~5.2]{zou2021} (Lower)&$\displaystyle\norm{\wb_0-\wb^*}_{\Hb_{k^\dagger:\infty}}^2\cdot\frac{k^*}{N}$\\[6pt]
\textbf{Corollary~\ref{cor:4.5}} (Ours, Upper \& Lower)&$\displaystyle\begin{aligned}&\frac{\norm{(\Ib-\gamma\Hb)^s(\wb_0-\wb^*)}_{\Ib_{0:k^*}}^2+N\gamma\norm{\wb_0-\wb^*}_{\Hb_{k^*:\infty}}^2}{N\gamma}\cdot\frac{k^*}{N}\\&\quad+\frac1{N^2}\sum_{i,j\le k^*}\frac{\lambda_j(\wb_0-\wb^*)_j^2(1-\gamma(\lambda_i\wedge\lambda_j))^{2(s-1)}}{s^{-1}\vee\gamma|\lambda_i-\lambda_j|}\end{aligned}$\\
\midrule
\textbf{Bound}&\textbf{Panel B: Slow-Mode Index Component ($i>k^*$)}\\
\midrule
\citet[Theorem~5.1]{zou2021} (Upper)&$\displaystyle\frac{\norm{\wb_0-\wb^*}_{\Ib_{0:k^\dagger}}^2+N\gamma\norm{\wb_0-\wb^*}_{\Hb_{k^\dagger:\infty}}^2}{N\gamma}\cdot\left(N\gamma^2\sum_{i>k^*}\lambda_i^2\right)$\\[6pt]
\citet[Theorem~5.2]{zou2021} (Lower)&$\displaystyle\norm{\wb_0-\wb^*}_{\Hb_{k^\dagger:\infty}}^2\cdot\left(N\gamma^2\sum_{i>k^*}\lambda_i^2\right)$\\[6pt]
\textbf{Corollary~\ref{cor:4.5}} (Ours, Upper \& Lower)&$\displaystyle\thupurple{\frac{\norm{\wb_0-\wb^*}_{\Ib_{0:k^\dagger}}^2+N\gamma\norm{\wb_0-\wb^*}_{\Hb_{k^\dagger:\infty}}^2}{N\gamma}\cdot\left(N\gamma^2\sum_{i>k^*}\lambda_i^2\right)}$\\
\bottomrule
\end{tabular}
\end{table}

\subsection{Sharp Separation Examples and Conditional Averaging-Window Bounds}\label{sec:4.4}
We conclude Section~\ref{sec:4} with two explicit separation constructions (Proposition~\ref{prop:4.7}) that demonstrate (i) an exponential-versus-polynomial gap between our lower bound in Theorem~\ref{thm:4.2} and \citet[Theorem~5.2]{zou2021}, and (ii) why fourth-moment assumptions alone cannot determine the asymptotic exponential burn-in rate without an additional structural condition (such as $\chi_{s,N}\le\chi_0<1$) or state information (such as $\Bb_s$ in Proposition~\ref{prop:4.8}).

\begin{proposition}[Sharp Separation Examples for Cross-Spectral Transfer and Burn-In Decay]\label{prop:4.7}
\leavevmode\par
\begin{enumerate}
\item \textup{(Exponential-vs.-Polynomial Gap in the Lower Bound of \citet{zou2021}):} Fix $0<\gamma\le1/6$ and integer $N\ge\max\{500,1/\gamma\}$. For $\xb\sim\cN(\zero,\Hb)$ in $\RR^2$ with $\Hb=\diag(1,(4\gamma N)^{-1})$, $y=\la\wb^*,\xb\ra$ ($\sigma_{\noise}^2=0$), $\wb_0-\wb^*=(1,0)^\top$, and $s=N$, constant-stepsize tail-averaged SGD satisfies
\begin{equation}\label{eq:4.17}
\EE\bigl[L(\overbar{\wb}_{N:2N})\bigr]-L(\wb^*)\ge\frac1{1024\gamma N^2},
\end{equation}
whereas \citet[Theorem~5.2]{zou2021} yields the exponentially small lower bound $(1-\gamma)^{2N}/(100\gamma^2N^2)$.
\item \textup{(Fourth Moments Shift Scalar Exponential Decay When $\chi_{s,N}$ Is Uncontrolled):} Fix $0<\gamma<1/3$ and $y=0$, $w^*=0$, $w_0=1$, $H=1$. Let $\cD_1$ be the three-point distribution $\PP(x=0)=1/2$, $\PP(x=\pm\sqrt2)=1/4$ (where $\EE[x^2]=1$, $\EE[x^4]=2$) and let $\cD_2$ be $\cN(0,1)$ (where $\EE[x^2]=1$, $\EE[x^4]=3$). Both satisfy Assumptions~\ref{ass:3.1}, \ref{ass:3.2}, and~\ref{ass:3.4} with $\alpha=3$ and $\beta=1$, yet their exact tail-averaged risks $R_\kappa(s,N)$ ($\kappa\in\{2,3\}$) satisfy $R_\kappa(s,N)\asymp_\gamma r_\kappa^s/N^2$ with $r_\kappa=1-2\gamma+\kappa\gamma^2$, so
\begin{equation}\label{eq:4.18}
\frac{R_3(N,N)}{R_2(N,N)}\gtrsim_\gamma\left(\frac{1-2\gamma+3\gamma^2}{1-2\gamma+2\gamma^2}\right)^N\longrightarrow\infty\qquad\text{as }N\to\infty.
\end{equation}
Consequently, no single expression depending only on $(\Hb,\wb_0-\wb^*,\gamma,s,N,\alpha,\beta,\sigma_{\noise}^2)$ can match both risks uniformly across $N$ without an additional structural condition (such as $\chi_{s,N}\le\chi_0<1$) or conditioning on $\Bb_s$.
\end{enumerate}
\end{proposition}

When $\chi_{s,N}\ge1$ (as in the large-$s$ scalar regime of Proposition~\ref{prop:4.7}(2)), conditioning at the start of the averaging window $t=s$ still yields upper and lower bounds in terms of the burn-in bias covariance $\Bb_s=\EE[\bmeta_s^{\bias}\otimes\bmeta_s^{\bias}]$ from Section~\ref{sec:3.3} that match up to constants depending only on $\alpha$, $\beta$, and $1-\gamma\alpha\tr(\Hb)$ for all $s\ge0$ and $N\ge2$ (recovering at the level of orders the result of restarting the full-averaging bounds of \citet{zou2021} at $\Bb_s$ while extending coverage to all $N\ge2$).

\begin{proposition}[Conditional Upper and Lower Bounds in Terms of $\Bb_s$]\label{prop:4.8}
Under Assumptions~\ref{ass:3.1}--\ref{ass:3.3} with $\gamma\le1/(2\alpha\tr(\Hb))$, for all $s\ge0$ and $N\ge1$, the excess risk satisfies
\begin{align}
\EE\bigl[L(\overbar{\wb}_{s:s+N})\bigr]-L(\wb^*)\le{}&\frac1{\gamma^2N^2}\tr\bigl(\Hb^{-1}_{0:k^*}\Bb_s\bigr)+\tr\bigl(\Hb_{k^*:\infty}\Bb_s\bigr)\notag\\
&+\frac{2\alpha}{1-\gamma\alpha\tr(\Hb)}\cdot\frac{\tr(\Ib_{0:k^*}\Bb_s)+N\gamma\tr(\Hb_{k^*:\infty}\Bb_s)}{N\gamma}\cdot\left(\frac{k^*}{N}+N\gamma^2\sum_{i>k^*}\lambda_i^2\right)\notag\\
&+\frac{2\sigma^2}{1-\gamma\alpha\tr(\Hb)}\cdot\left(\frac{k^*}{N}+\gamma\sum_{k^*<i\le k^\dagger}\lambda_i+(s+N)\gamma^2\sum_{i>k^\dagger}\lambda_i^2\right),\label{eq:4.19}
\end{align}
and under \eqref{eq:3.4} and Assumptions~\ref{ass:3.1}, \ref{ass:3.2}, and~\ref{ass:3.4} with $\gamma\le1/(2\lambda_1)$ and $N\ge2$,
\begin{align}
\EE\bigl[L(\overbar{\wb}_{s:s+N})\bigr]-L(\wb^*)\ge{}&\frac1{8\gamma^2N^2}\tr\bigl(\Hb^{-1}_{0:k^*}\Bb_s\bigr)+\frac18\tr\bigl(\Hb_{k^*:\infty}\Bb_s\bigr)\notag\\
&+\frac\beta{864}\cdot\frac{\tr(\Ib_{0:k^*}\Bb_s)+N\gamma\tr(\Hb_{k^*:\infty}\Bb_s)}{N\gamma}\cdot\left(\frac{k^*}{N}+N\gamma^2\sum_{i>k^*}\lambda_i^2\right)\notag\\
&+\frac{\sigma_{\noise}^2}{216}\cdot\left(\frac{k^*}{N}+\gamma\sum_{k^*<i\le k^\dagger}\lambda_i+(s+N)\gamma^2\sum_{i>k^\dagger}\lambda_i^2\right).\label{eq:4.20}
\end{align}
\end{proposition}

\section{Overview of Proof Techniques}\label{sec:5}
In this section, we present the analytical framework underlying Theorems~\ref{thm:4.1} and~\ref{thm:4.2}, Proposition~\ref{prop:4.3}, Corollary~\ref{cor:4.4}, and Propositions~\ref{prop:4.7}--\ref{prop:4.8}. Recall from Section~\ref{sec:3.3} the bias and variance error processes $\{\bmeta_t^{\bias}\}_{t\ge0}$ and $\{\bmeta_t^{\variance}\}_{t\ge0}$ in \eqref{eq:5.1}--\eqref{eq:5.2} and their single-step second-moment operators $\Bb_t=\EE[\bmeta_t^{\bias}\otimes\bmeta_t^{\bias}]$ and $\Cb_t=\EE[\bmeta_t^{\variance}\otimes\bmeta_t^{\variance}]$. Following \citet{jain2017b} and \citet{zou2021}, we define the linear transition super-operators $\cT$ and $\wt{\cT}$ acting on symmetric operators $\Ab$ by
\begin{equation}
\cT:=\Hb\otimes\Ib+\Ib\otimes\Hb-\gamma\cM,\qquad
\wt{\cT}:=\Hb\otimes\Ib+\Ib\otimes\Hb-\gamma\wt{\cM},
\label{eq:5.3}
\end{equation}
so that $(\cI-\gamma\cT)\circ\Ab=\EE[(\Ib-\gamma\xb\xb^\top)\Ab(\Ib-\gamma\xb\xb^\top)]$ is the single-step stochastic covariance transition and $(\cI-\gamma\wt{\cT})\circ\Ab=(\Ib-\gamma\Hb)\Ab(\Ib-\gamma\Hb)$ is its deterministic counterpart. Their difference isolates the fourth-moment fluctuation super-operator:
\begin{equation}
(\cI-\gamma\cT)-(\cI-\gamma\wt{\cT})=\gamma^2(\cM-\wt{\cM}).\label{eq:5.4}
\end{equation}
Let $\cF_t:=\sigma(\{(\xb_u,y_u)\}_{u=1}^t)$ denote the natural filtration generated by the first $t$ samples (with $\cF_0$ trivial). Because $(\xb_t,y_t)$ is independent of $\cF_{t-1}$ and first-order optimality implies $\EE[\xi_t\xb_t]=\nabla L(\wb^*)=\zero$, taking expectations in \eqref{eq:5.2} from $\bmeta_0^{\variance}=\zero$ gives $\EE[\bmeta_t^{\variance}]=(\Ib-\gamma\Hb)\EE[\bmeta_{t-1}^{\variance}]=\zero$ for all $t\ge1$. Together with sample independence, this cancels the cross terms in $\Cb_t$, so taking outer products and expectations in \eqref{eq:5.1}--\eqref{eq:5.2} yields the exact single-step operator recurrences \citep{jain2017b,zou2021}:
\begin{equation}
\Bb_t=(\cI-\gamma\cT)\circ\Bb_{t-1},\qquad
\Cb_t=(\cI-\gamma\cT)\circ\Cb_{t-1}+\gamma^2\bSigma,\qquad \forall t\geq1.
\label{eq:5.5}
\end{equation}

Our analysis is organized into two main components. In Section~\ref{sec:5.1}, we develop a perturbation-theoretic trajectory decomposition (Lemma~\ref{lem:5.1}) that unrolls \eqref{eq:5.4} across the full horizon and leads directly to a discrete Volterra feedback absorption inequality (Lemma~\ref{lem:5.2}, whose complete proof we present below). In Section~\ref{sec:5.2}, we analyze the resulting propagation kernel $k(h)$ and explain how the pair-midpoint convolution identity and kernel doubling constant $D_T$ control $\chi_{s,N}$ across general and polynomially decaying spectra.

\subsection{Perturbation Decomposition and Discrete Volterra Feedback Absorption}\label{sec:5.1}
In \citet{zou2021}, the tail-averaged bias error is analyzed by splitting the trajectory at the end of the burn-in phase $t=s$: one first unrolls \eqref{eq:5.5} over $0\le t\le s-1$ via the Duhamel expansion
\begin{equation}
\Bb_s=(\Ib-\gamma\Hb)^s\Bb_0(\Ib-\gamma\Hb)^s+\gamma^2\sum_{t=0}^{s-1}(\Ib-\gamma\Hb)^{s-1-t}\bigl((\cM-\wt{\cM})\circ\Bb_t\bigr)(\Ib-\gamma\Hb)^{s-1-t},
\label{eq:5.7}
\end{equation}
and then restarts a full-averaging bound from $\Bb_s$ over $[s,s+N-1]$. The subsequent relaxations in this two-stage argument cause both the upper- and lower-bound gaps in \eqref{eq:1.1}--\eqref{eq:1.2}: \citet[Lemma~D.4]{zou2021} upper-bounds \eqref{eq:5.7} by replacing $(\Ib-\gamma\Hb)^{2(s-1-t)}\Hb$ with $\Hb$ (and \citet[Lemma~D.5]{zou2021} replaces the averaging-window trace drop $\tr(\Bb_{s,s+N})$ by $\tr(\Bb_{0,s+N})$, where $\Bb_{a,b}:=\Bb_a-(\Ib-\gamma\Hb)^{b-a}\Bb_a(\Ib-\gamma\Hb)^{b-a}$), which destroys the remaining burn-in contraction on the fast modes; conversely, \citet[Lemma~D.9]{zou2021} lower-bounds \eqref{eq:5.7} by dropping the non-negative sum over $t\in\{0,\ldots,s-1\}$ altogether, which discards all burn-in fourth-moment energy scattered from the head initialization into slow modes.

Inspired by the operator-polynomial reformulations developed for momentum and exponential moving average SGD in \citet{li2023,li2025}, we bypass the intermediate boundary $t=s$ and unroll the perturbation identity \eqref{eq:5.4} directly across the entire trajectory $r\in\{0,\ldots,s+N-2\}$. Physically, a stochastic gradient fluctuation injected at step $t=r+1$ is propagated by the deterministic transition operator $(\Ib-\gamma\Hb)^{m-t}$ to every subsequent iterate $m\in\{s\vee t,\ldots,T\}$ in the averaging window, accumulating the filter polynomial $\Qb_{r,s,N}=q_t(\Hb)$ from \eqref{eq:4.2}. This yields the following exact covariance decomposition, proved in Appendix~\ref{app:A.1}.

\begin{lemma}[Exact Operator-Polynomial Decomposition for Tail-Averaging]\label{lem:5.1}
Under Assumptions~\ref{ass:3.1}--\ref{ass:3.3}, for any stepsize $\gamma>0$, burn-in length $s\geq0$, and averaging window $N\geq1$, define the deterministic operator polynomials
\begin{equation}
\Pb_{s,N}:=\frac1N\sum_{t=s}^{s+N-1}(\Ib-\gamma\Hb)^t,\qquad
\Qb_{r,s,N}:=\frac1N\sum_{t=s\vee(r+1)}^{s+N-1}(\Ib-\gamma\Hb)^{t-1-r},\quad \forall r\in\{0,\ldots,s+N-2\}.
\label{eq:5.8}
\end{equation}
Then the tail-averaged bias and variance covariance operators satisfy the exact identities
\begin{align}
\EE\bigl[\overbar{\bmeta}_{s:s+N}^{\bias}\otimes\overbar{\bmeta}_{s:s+N}^{\bias}\bigr]
&=\Pb_{s,N}\Bb_0\Pb_{s,N}+\gamma^2\sum_{r=0}^{s+N-2}\Qb_{r,s,N}\bigl((\cM-\wt{\cM})\circ\Bb_r\bigr)\Qb_{r,s,N},\label{eq:5.9}\\
\EE\bigl[\overbar{\bmeta}_{s:s+N}^{\variance}\otimes\overbar{\bmeta}_{s:s+N}^{\variance}\bigr]
&=\gamma^2\sum_{r=0}^{s+N-2}\Qb_{r,s,N}\bigl((\cM-\wt{\cM})\circ\Cb_r+\bSigma\bigr)\Qb_{r,s,N}.\label{eq:5.10}
\end{align}
\end{lemma}

Lemma~\ref{lem:5.1} explains why the receiving weights $F_t$, propagation kernel $k(h)$, and spectral feedback ratio $\chi_{s,N}$ in \eqref{eq:4.2}--\eqref{eq:4.3} arise from first principles. Taking the trace inner product of \eqref{eq:5.9} and \eqref{eq:5.10} with $\frac12\Hb$, the deterministic term $\frac12\la\Hb,\Pb_{s,N}\Bb_0\Pb_{s,N}\ra$ equals $\EffectiveBias$ in \eqref{eq:4.1}, while the remaining stochastic terms sum to $\EffectiveVar=(\bias-\EffectiveBias)+\variance$ in \eqref{eq:effvar_def}. Under Assumptions~\ref{ass:3.2}--\ref{ass:3.4}, the fourth-moment fluctuation operators $(\cM-\wt{\cM})\circ\Bb_{t-1}$ and $(\cM-\wt{\cM})\circ\Cb_{t-1}$ are sandwiched by multiples of $\tr(\Hb\Bb_{t-1})\Hb$ and $\tr(\Hb\Cb_{t-1})\Hb$. Consequently, each fluctuation excited along $\Hb$ at step $t=r+1\in\{1,\ldots,T\}$ is weighted at the output by
\[
\inner{\Hb}{\Qb_{t-1,s,N}\Hb\Qb_{t-1,s,N}}=\tr\bigl(\Hb^2q_t(\Hb)^2\bigr)=\sum_{i=1}^{\infty}\lambda_i^2q_t(\lambda_i)^2=F_t.
\]
When $\Bb_{t-1}$ is replaced by its unperturbed deterministic drift $(\Ib-\gamma\Hb)^{t-1}\Bb_0(\Ib-\gamma\Hb)^{t-1}$, the resulting single-hop initialization-transfer functional $\cS_{s,N}$ and additive-noise functional $\cV_{s,N}$ take the form
\begin{equation}
\cS_{s,N}:=\frac{\gamma^2}{2}\sum_{t=1}^T\norm{(\Ib-\gamma\Hb)^{t-1}(\wb_0-\wb^*)}_{\Hb}^2F_t,\qquad
\cV_{s,N}:=\frac{\gamma^2}{2}\sum_{t=1}^TF_t.
\label{eq:5.11}
\end{equation}

To bound the multi-hop contributions in $\EffectiveVar$, we must control how the covariance traces $f_r:=\tr(\Hb\Bb_r)$ and $g_r:=\tr(\Hb\Cb_r)$ amplify over time. Unrolling $\Bb_{t-1}$ as in \eqref{eq:5.7} (and $\Cb_{t-1}$ analogously) and taking the trace inner product with $\Hb$ reveals that a fluctuation of magnitude $f_{r-1}\Hb$ injected at an earlier step $r<t$ is contracted by $(\Ib-\gamma\Hb)^{t-r-1}$ before being tested against $\Hb$ at step $t-1$, producing the two-time trace propagation kernel
\[
\tr\bigl(\Hb(\Ib-\gamma\Hb)^{t-r-1}\Hb(\Ib-\gamma\Hb)^{t-r-1}\bigr)=\sum_{i=1}^{\infty}\lambda_i^2(1-\gamma\lambda_i)^{2(t-r-1)}=k(t-r-1).
\]
Thus $\{f_t\}$ and $\{g_t\}$ obey a discrete-time Volterra convolution recursion driven by the kernel $k(t-r-1)$---a non-asymptotic counterpart of the Volterra integral equations that govern high-dimensional SGD dynamics \citep{paquette2021sgd,paquette2022halting,li2025functional}. Crucially, because the receiving weights $F_t$ \emph{increase} with $t$ during the burn-in phase $1\le t\le s$, bounding $\sum_{t=1}^T F_t f_{t-1}$ via an unweighted trace sum $\sum_{t=1}^T f_{t-1}$ would require replacing $F_t$ by $\max_t F_t$ and losing burn-in contraction. Inspired by the Volterra convolution self-bounding strategy of \citet[Appendix~C.3.1]{li2025functional}, we instead bound the $F_t$-weighted trajectory sum directly as a self-bounding functional.

\begin{lemma}[Weighted Trajectory Feedback Absorption]\label{lem:5.2}
Suppose Assumption~\ref{ass:3.1} holds and $0<\gamma<1/\lambda_1$, and recall $\EffectiveBias$, $\EffectiveVar$, $\chi_{s,N}$, $\cS_{s,N}$, and $\cV_{s,N}$ from \eqref{eq:4.1}, \eqref{eq:effvar_def}, \eqref{eq:4.3}, and \eqref{eq:5.11}.
\begin{enumerate}
\item Under Assumptions~\ref{ass:3.2} and~\ref{ass:3.3}, if $\chi_{s,N}<1$, then
\begin{equation}
\EffectiveVar\leq\frac{\alpha}{1-\chi_{s,N}}\cS_{s,N}+\frac{\sigma^2}{1-\chi_{s,N}}\cV_{s,N}.
\label{eq:5.12}
\end{equation}
\item Under the well-specified model \eqref{eq:3.4} and Assumptions~\ref{ass:3.2} and~\ref{ass:3.4} (without requiring $\chi_{s,N}<1$),
\begin{equation}
\EffectiveVar\geq\beta\cS_{s,N}+\sigma_{\noise}^2\cV_{s,N}.
\label{eq:5.13}
\end{equation}
\end{enumerate}
\end{lemma}

\begin{proof}[Proof of Lemma~\ref{lem:5.2}]
For each integer $r\ge0$, let $f_r:=\tr(\Hb\Bb_r)$, $g_r:=\tr(\Hb\Cb_r)$, and $b_r:=\tr(\Hb(\Ib-\gamma\Hb)^{2r}\Bb_0)=\norm{(\Ib-\gamma\Hb)^r(\wb_0-\wb^*)}_{\Hb}^2$. Taking the trace inner product of \eqref{eq:5.9} and \eqref{eq:5.10} in Lemma~\ref{lem:5.1} with $\frac12\Hb$ and using $\frac12\la\Hb,\Pb_{s,N}\Bb_0\Pb_{s,N}\ra=\EffectiveBias$ together with the index shift $t=r+1\in\{1,\ldots,T\}$, we obtain the exact representation
\begin{equation}\label{eq:5.14}
\EffectiveVar=\frac{\gamma^2}{2}\sum_{t=1}^{T}\inner{\Hb}{\Qb_{t-1,s,N}\Bigl((\cM-\wt{\cM})\circ(\Bb_{t-1}+\Cb_{t-1})+\bSigma\Bigr)\Qb_{t-1,s,N}}.
\end{equation}

We first prove the upper bound \eqref{eq:5.12}. Applying Assumptions~\ref{ass:3.2} and~\ref{ass:3.3} to the inner operator in \eqref{eq:5.14} gives $(\cM-\wt{\cM})\circ\Bb_{t-1}\preceq\alpha f_{t-1}\Hb$ and $(\cM-\wt{\cM})\circ\Cb_{t-1}+\bSigma\preceq(\alpha g_{t-1}+\sigma^2)\Hb$. Substituting these Loewner upper bounds into \eqref{eq:5.14} yields
\begin{align}
\EffectiveVar
&\le\frac{\gamma^2}{2}\sum_{t=1}^{T}\bigl[\alpha(f_{t-1}+g_{t-1})+\sigma^2\bigr]\inner{\Hb}{\Qb_{t-1,s,N}\Hb\Qb_{t-1,s,N}}\notag\\
&=\frac{\gamma^2}{2}\sum_{t=1}^{T}F_t\bigl[\alpha(f_{t-1}+g_{t-1})+\sigma^2\bigr]=:U,\label{eq:5.15}
\end{align}
where the equality uses $\la\Hb,\Qb_{t-1,s,N}\Hb\Qb_{t-1,s,N}\ra=F_t$. Next, unrolling the single-step recurrences \eqref{eq:5.5} via the perturbation identity \eqref{eq:5.4} up to step $t-1$ gives
\begin{align*}
\Bb_{t-1}&=(\Ib-\gamma\Hb)^{t-1}\Bb_0(\Ib-\gamma\Hb)^{t-1}+\gamma^2\sum_{r=1}^{t-1}(\Ib-\gamma\Hb)^{t-1-r}\bigl((\cM-\wt{\cM})\circ\Bb_{r-1}\bigr)(\Ib-\gamma\Hb)^{t-1-r},\\
\Cb_{t-1}&=\gamma^2\sum_{r=1}^{t-1}(\Ib-\gamma\Hb)^{t-1-r}\bigl((\cM-\wt{\cM})\circ\Cb_{r-1}+\bSigma\bigr)(\Ib-\gamma\Hb)^{t-1-r}.
\end{align*}
Taking the trace inner product of both expansions with $\Hb$ and applying Assumptions~\ref{ass:3.2} and~\ref{ass:3.3} together with $\tr(\Hb(\Ib-\gamma\Hb)^{t-1-r}\Hb(\Ib-\gamma\Hb)^{t-1-r})=k(t-r-1)$, we obtain for every $t\in\{1,\ldots,T\}$ the coupled Volterra trace inequalities
\begin{equation}\label{eq:5.16}
f_{t-1}\le b_{t-1}+\alpha\gamma^2\sum_{r=1}^{t-1}k(t-r-1)f_{r-1},\qquad
g_{t-1}\le\gamma^2\sum_{r=1}^{t-1}k(t-r-1)\bigl(\alpha g_{r-1}+\sigma^2\bigr).
\end{equation}
Substituting \eqref{eq:5.16} into the definition of $U$ in \eqref{eq:5.15} yields
\begin{align}
U&\le\frac{\gamma^2}{2}\sum_{t=1}^{T}F_t\bigl(\alpha b_{t-1}+\sigma^2\bigr)
+\frac{\gamma^2}{2}\sum_{t=1}^{T}F_t\alpha\gamma^2\sum_{r=1}^{t-1}k(t-r-1)\bigl[\alpha(f_{r-1}+g_{r-1})+\sigma^2\bigr]\notag\\
&=\alpha\cS_{s,N}+\sigma^2\cV_{s,N}
+\frac{\gamma^2}{2}\sum_{r=1}^{T-1}\bigl[\alpha(f_{r-1}+g_{r-1})+\sigma^2\bigr]\left(\alpha\gamma^2\sum_{t=r+1}^{T}k(t-r-1)F_t\right)\notag\\
&\le\alpha\cS_{s,N}+\sigma^2\cV_{s,N}
+\chi_{s,N}\cdot\frac{\gamma^2}{2}\sum_{r=1}^{T-1}F_r\bigl[\alpha(f_{r-1}+g_{r-1})+\sigma^2\bigr]\notag\\
&\le\alpha\cS_{s,N}+\sigma^2\cV_{s,N}+\chi_{s,N}U,\label{eq:5.17}
\end{align}
where the first equality uses the definitions of $\cS_{s,N}$ and $\cV_{s,N}$ in \eqref{eq:5.11} and interchanges the finite sums over $1\le r<t\le T$, the second inequality uses $\alpha\gamma^2\sum_{t=r+1}^{T}k(t-r-1)F_t\le\chi_{s,N}F_r$ from the definition of $\chi_{s,N}$ in \eqref{eq:4.3}, and the last inequality extends the non-negative sum over $r$ to $T$. Since $\chi_{s,N}<1$, rearranging \eqref{eq:5.17} and combining with \eqref{eq:5.15} establishes \eqref{eq:5.12}.

Finally, to prove the lower bound \eqref{eq:5.13}, Lemma~\ref{lem:C.2}(3) implies $\Bb_{t-1}\succeq(\Ib-\gamma\Hb)^{t-1}\Bb_0(\Ib-\gamma\Hb)^{t-1}$ (so $f_{t-1}\ge b_{t-1}$) and $(\cM-\wt{\cM})\circ\Cb_{t-1}\succeq\zero$. Applying Assumption~\ref{ass:3.4} and $\bSigma=\sigma_{\noise}^2\Hb$ under \eqref{eq:3.4} to \eqref{eq:5.14} gives
\begin{align*}
\EffectiveVar
&\ge\frac{\gamma^2}{2}\sum_{t=1}^{T}\inner{\Hb}{\Qb_{t-1,s,N}\bigl(\beta f_{t-1}\Hb+\sigma_{\noise}^2\Hb\bigr)\Qb_{t-1,s,N}}\\
&\ge\frac{\gamma^2}{2}\sum_{t=1}^{T}F_t\bigl(\beta b_{t-1}+\sigma_{\noise}^2\bigr)
=\beta\cS_{s,N}+\sigma_{\noise}^2\cV_{s,N},
\end{align*}
which completes the proof of Lemma~\ref{lem:5.2}.
\end{proof}

\subsection{Propagation Kernel Geometry, Pair-Midpoint Convolution, and Spectral Regimes}\label{sec:5.2}
With Lemma~\ref{lem:5.2} in hand, the remaining analytical challenge is to bound the spectral feedback ratio $\chi_{s,N}$ in \eqref{eq:4.3} without losing the denominator weight $F_r$. During the burn-in phase $1\le r<t\le s$, the receiving sensitivity $F_t$ increases with $t$ because later perturbations undergo fewer burn-in contractions before entering the averaging window $[s,T]$. Consequently, bounding $F_t/F_r$ pointwise would incur an exponential blow-up in $s$.

Our key observation (Proposition~\ref{prop:4.3}) is that $F_t$ and the step-to-step kernel $k(h)$ are governed by the \emph{same} functional $k(\cdot)$. Expanding the square of the filter polynomial $q_t(\lambda_i)=\frac1N\sum_{a=s\vee t}^T(1-\gamma\lambda_i)^{a-t}$ in \eqref{eq:4.2} and summing against $\lambda_i^2$ expresses $F_t$ as the average of $k(\cdot)$ evaluated at the pair midpoint $\frac{a+b}{2}-t$:
\[
F_t=\sum_{i=1}^{\infty}\lambda_i^2\left(\frac1N\sum_{a=s\vee t}^{T}(1-\gamma\lambda_i)^{a-t}\right)\left(\frac1N\sum_{b=s\vee t}^{T}(1-\gamma\lambda_i)^{b-t}\right)
=\frac1{N^2}\sum_{a,b=s\vee t}^{T}k\Bigl(\frac{a+b}{2}-t\Bigr).
\]
Substituting this pair-midpoint representation into $\gamma^2\sum_{t=r+1}^T k(t-r-1)F_t$ and setting $j:=t-r-1$ and $h:=\frac{a+b}{2}-r\in[1,T]$ reduces the inner sum over $t$ to the discrete self-convolution $\sum_{j=0}^{\min(a,b)-r-1}k(j)k(h-1-j)$ at total lag $h-1$. This structure yields two complementary mechanisms for controlling $\chi_{s,N}$ (proved in Appendix~\ref{app:B.2}):
\begin{itemize}
\item \emph{General Trace-Class Spectra via Mode-Wise Positive Correlation (Proposition~\ref{prop:4.3}(1)):}
Under the main-theorem restriction $\gamma\lambda_1\le1/2$, write $a_i:=1-\gamma\lambda_i\in[1/2,1)$. Because slower modes contract more slowly, the sequences $\{a_i^{2j}\}_{i\ge1}$ and $\{a_i^{2(h-1-j)}\}_{i\ge1}$ are both non-decreasing in $i$. By Chebyshev's sum inequality under the positive weights $\{\lambda_i^2\}_{i\ge1}$ (Lemma~\ref{lem:C.0}),
\[
k(j)k(h-1-j)\le\tr(\Hb^2)k(h-1)\le\frac{\tr(\Hb^2)}{(1-\gamma\lambda_1)^2}k\Bigl(\frac{a+b}{2}-r\Bigr)\lesssim\tr(\Hb^2)k\Bigl(\frac{a+b}{2}-r\Bigr).
\]
Summing over the at most $T-1$ indices $j$ and averaging over $a,b\ge s\vee(r+1)\ge s\vee r$ reconstructs $F_r=\frac1{N^2}\sum_{a,b=s\vee r}^T k(\frac{a+b}{2}-r)$, canceling the denominator $F_r$ in \eqref{eq:4.3} and yielding $\chi_{s,N}\lesssim\alpha\gamma^2 T\tr(\Hb^2)$.

\item \emph{Polynomial Spectra via Midpoint Convolution Splitting and Kernel Doubling (Proposition~\ref{prop:4.3}(2) and Corollary~\ref{cor:4.4}):}
In every term of the self-convolution $\sum_{j=0}^{h-1}k(j)k(h-1-j)$, at least one of the two lags $j$ or $h-1-j$ is at least $(h-1)/2$. Since $k(\cdot)$ is non-increasing, splitting the convolution at the midpoint $(h-1)/2$ (a discrete counterpart of regular-variation splitting for power-law Volterra kernels; cf.\ \citet{li2025functional}) factors out $k((h-1)/2)\le D_T k(h)$ while the remaining factor sums to $\gamma^2\sum_{n=0}^\infty k(n)\asymp\gamma\tr(\Hb)$:
\[
\gamma^2\sum_{j=0}^{\min(a,b)-r-1}k(j)k(h-1-j)
\le2k\Bigl(\frac{h-1}{2}\Bigr)\gamma^2\sum_{n=0}^{\infty}k(n)
\lesssim D_T\gamma\tr(\Hb)\,k\Bigl(\frac{a+b}{2}-r\Bigr).
\]
Averaging over $(a,b)$ again recovers $F_r$ and gives the horizon-free bound $\chi_{s,N}\lesssim\alpha D_T\gamma\tr(\Hb)$. Under two-sided polynomial eigenvalue decay $\lambda_i\asymp\Lambda i^{-a}$ ($a>1$), integral comparison over the active spectral scale $i\asymp(\gamma\Lambda h)^{1/a}$ shows that $k(h)\asymp\Lambda^2(1+\gamma\Lambda h)^{-(2-1/a)}$ decays as a power law, so its doubling ratio $D_T$ is uniformly bounded by a constant $D_*(a,\theta)<\infty$ for all $T\ge1$ (whereas in $d=1$, $k(h)=\lambda_1^2(1-\gamma\lambda_1)^{2h}$ decays exponentially so $D_T$ grows exponentially in $T$, giving rise to the scalar separation in Proposition~\ref{prop:4.7}(2)).
\end{itemize}
Finally, in Appendix~\ref{app:A.2} (Lemma~\ref{lem:A.3}), we evaluate the single-hop functionals $\EffectiveBias$, $\cV_{s,N}$, and $\cS_{s,N}$ within universal constants across the spectral cutoffs $k^*$ and $k^\dagger$, which combined with Lemma~\ref{lem:5.2} yields Theorems~\ref{thm:4.1} and~\ref{thm:4.2}.

\section{Conclusion}\label{sec:6}
We have established initialization-explicit matching upper and lower excess risk bounds for constant-stepsize stochastic gradient descent with tail-averaging in high-dimensional linear regression under the distributional assumptions of \citet{zou2021}. By controlling cumulative fourth-moment trajectory feedback through the spectral feedback ratio $\chi_{s,N}\le\chi_0<1$ and bounding $\chi_{s,N}$ via the pair-midpoint convolution identity, we proved that matching bounds hold for arbitrary covariance spectra under a horizon-dependent constant stepsize chosen before the run, and hold uniformly across all horizons $s\ge0$ and $N\ge2$ at a single horizon-independent stepsize for polynomially decaying spectra $\lambda_i\asymp i^{-a}$ ($a>1$). Complementing these positive results, our two-dimensional Gaussian separation construction demonstrates an exponential-versus-polynomial improvement over the lower bound of \citet[Theorem~5.2]{zou2021} due to burn-in head-to-tail transfer, while our scalar separation construction shows why additional structural conditions (such as $\chi_{s,N}\le\chi_0<1$) or state information ($\Bb_s$) are required over the unrestricted fourth-moment class when $D_T$ grows exponentially.

\section*{Acknowledgment}

The proof ideas are developed through an interactive brainstorm among GPT-6 Astra, Gemini 4 Argon, and the authors. Gemini 4 Argon are employed to further streamline the canonical pillars underlying the proof roadmap. The authors simplified the arguments, re-organized the proofs, verified the details, and prepared the draft with further assistance from Gemini 4 Argon and GPT-6 Astra. The authors take full responsibility for its contents.

\clearpage
\appendix
\section{Trajectory Decomposition and Spectral Evaluation}\label{app:A}
In this appendix, we establish the exact operator-polynomial decomposition (Lemma~\ref{lem:5.1} in Section~\ref{app:A.1}) and the two-sided spectral evaluations of $\EffectiveBias$, $\cV_{s,N}$, and $\cS_{s,N}$ (Lemma~\ref{lem:A.3} in Section~\ref{app:A.2}). Elementary algebraic and convolution lemmas as well as foundational operator lemmas from prior work are collected in Appendix~\ref{app:C}.

\paragraph{Additional Notations.}
Throughout the appendices, for any real number $z\in\RR$, we write $\floor{z}$ and $\ceil{z}$ for the floor and ceiling of $z$, respectively. Recall from Section~\ref{sec:5} the natural filtration $\cF_t=\sigma(\{(\xb_u,y_u)\}_{u=1}^t)$ (with $\cF_0$ trivial) and the eigenvalue cutoffs $k^*=\max\{k:\lambda_k\ge(\gamma N)^{-1}\}$ and $k^\dagger=\max\{k:\lambda_k\ge(\gamma(s+N))^{-1}\}$ from \eqref{eq:4.6}. To streamline the spectral evaluations across Appendices~\ref{app:A} and~\ref{app:B}, we define the burn-in drifted error vector, initial eigen-coordinates, and window effective-variance multiplier by
\begin{equation}\label{eq:A.0}
\wb_s^{\drift}:=(\Ib-\gamma\Hb)^s(\wb_0-\wb^*),\qquad
\omega_j:=(\wb_0-\wb^*)_j=\la\vb_j,\wb_0-\wb^*\ra,\qquad
\Xi_N:=\frac{k^*}{N}+N\gamma^2\sum_{i>k^*}\lambda_i^2,
\end{equation}
and for each $r\in\{0,\ldots,s+N-2\}$, we define the step-$(r+1)$ scaled receiving weight
\begin{equation}\label{eq:A.0b}
\Psi_{r,s,N}:=\gamma^2F_{r+1}
=\frac1{N^2}\sum_{i=1}^{\infty}(1-\gamma\lambda_i)^{2((s-1-r)\vee0)}\bigl(1-(1-\gamma\lambda_i)^{(s+N-1-r)\wedge N}\bigr)^2.
\end{equation}

\subsection{Proof of Lemma \ref{lem:5.1}}\label{app:A.1}
\begin{proof}[Proof of Lemma~\ref{lem:5.1}]
We first establish the bias covariance identity~\eqref{eq:5.9}. For each $r\ge1$, define the zero-mean random operator fluctuation
\[
\bDelta_r:=(\Ib-\gamma\xb_r\xb_r^\top)-(\Ib-\gamma\Hb)=\gamma(\Hb-\xb_r\xb_r^\top).
\]
Then the single-step bias recursion~\eqref{eq:5.1} can be written as $\bmeta_t^{\bias}=(\Ib-\gamma\Hb)\bmeta_{t-1}^{\bias}+\bDelta_t\bmeta_{t-1}^{\bias}$. Unrolling this linear recursion from step $0$ to step $t\ge0$ gives
\begin{equation}\label{eq:A.1}
\bmeta_t^{\bias}=(\Ib-\gamma\Hb)^t\bmeta_0^{\bias}+\sum_{r=0}^{t-1}(\Ib-\gamma\Hb)^{t-1-r}\bDelta_{r+1}\bmeta_r^{\bias}.
\end{equation}
Averaging~\eqref{eq:A.1} over $t\in\{s,s+1,\ldots,s+N-1\}$ and exchanging the order of summation between $t$ and $r\in\{0,\ldots,s+N-2\}$, we obtain
\begin{align}
\overbar{\bmeta}_{s:s+N}^{\bias}
&=\frac1N\sum_{t=s}^{s+N-1}(\Ib-\gamma\Hb)^t\bmeta_0^{\bias}
 +\frac1N\sum_{t=s}^{s+N-1}\sum_{r=0}^{t-1}(\Ib-\gamma\Hb)^{t-1-r}\bDelta_{r+1}\bmeta_r^{\bias}\notag\\
&=\Pb_{s,N}\bmeta_0^{\bias}+\sum_{r=0}^{s+N-2}\left(\frac1N\sum_{t=s\vee(r+1)}^{s+N-1}(\Ib-\gamma\Hb)^{t-1-r}\right)\bDelta_{r+1}\bmeta_r^{\bias}\notag\\
&=\Pb_{s,N}\bmeta_0^{\bias}+\sum_{r=0}^{s+N-2}\Qb_{r,s,N}\bDelta_{r+1}\bmeta_r^{\bias},\label{eq:A.2}
\end{align}
where the third equality uses the definitions of $\Pb_{s,N}$ and $\Qb_{r,s,N}$ in~\eqref{eq:5.8}. Since $\bmeta_r^{\bias}$ is $\cF_r$-measurable and $(\xb_{r+1},y_{r+1})$ is independent of $\cF_r$ with $\EE[\bDelta_{r+1}\mid\cF_r]=\zero$, the sequence $\{\Qb_{r,s,N}\bDelta_{r+1}\bmeta_r^{\bias}\}_{r=0}^{s+N-2}$ is a martingale difference sequence with respect to $\{\cF_{r+1}\}_{r\ge0}$. Consequently, for any $0\le r<r'\le s+N-2$,
\[
\EE\big[(\Qb_{r',s,N}\bDelta_{r'+1}\bmeta_{r'}^{\bias})\otimes(\Qb_{r,s,N}\bDelta_{r+1}\bmeta_r^{\bias})\big]=\zero,
\qquad
\EE\big[(\Qb_{r,s,N}\bDelta_{r+1}\bmeta_r^{\bias})\otimes(\Pb_{s,N}\bmeta_0^{\bias})\big]=\zero.
\]
Furthermore, for each $r\in\{0,\ldots,s+N-2\}$, conditioning on $\cF_r$ and using~\eqref{eq:3.3} gives
\begin{align}
\EE\big[(\bDelta_{r+1}\bmeta_r^{\bias})\otimes(\bDelta_{r+1}\bmeta_r^{\bias})\big]
&=\gamma^2\EE\Big[\EE\big[(\xb_{r+1}\xb_{r+1}^\top-\Hb)(\bmeta_r^{\bias}\otimes\bmeta_r^{\bias})(\xb_{r+1}\xb_{r+1}^\top-\Hb)\mid\cF_r\big]\Big]\notag\\
&=\gamma^2(\cM-\wt\cM)\circ\EE\big[\bmeta_r^{\bias}\otimes\bmeta_r^{\bias}\big]=\gamma^2(\cM-\wt\cM)\circ\Bb_r.\label{eq:A.3}
\end{align}
Taking the outer product of~\eqref{eq:A.2} with itself in expectation and substituting~\eqref{eq:A.3} yields~\eqref{eq:5.9}.

To prove the variance covariance identity~\eqref{eq:5.10} under Assumption~\ref{ass:3.3} (without assuming independence of $\xi_t$ and $\xb_t$), we work with $\Cb_t=\EE[\bmeta_t^{\variance}\otimes\bmeta_t^{\variance}]$. Since $\bmeta_t^{\variance}$ is $\cF_t$-measurable and the tower property with $\EE[\xi_u\xb_u\mid\cF_{u-1}]=\nabla L(\wb^*)=\zero$ implies $\EE[\xi_u\xb_u\mid\cF_t]=\zero$ for all $u>t$, taking conditional expectations given $\cF_t$ in~\eqref{eq:5.2} yields $\EE[\bmeta_{t'}^{\variance}\mid\cF_t]=(\Ib-\gamma\Hb)^{t'-t}\bmeta_t^{\variance}$ for all $t'\ge t\ge0$, and hence
\begin{equation}\label{eq:A.4}
\EE\big[\bmeta_{t'}^{\variance}\otimes\bmeta_t^{\variance}\big]
=\EE\Big[\EE\big[\bmeta_{t'}^{\variance}\mid\cF_t\big]\otimes\bmeta_t^{\variance}\Big]
=(\Ib-\gamma\Hb)^{t'-t}\Cb_t,\qquad\forall t'\ge t\ge0.
\end{equation}
On the other hand, iterating $\Cb_t=(\cI-\gamma\wt\cT)\circ\Cb_{t-1}+\gamma^2\big((\cM-\wt\cM)\circ\Cb_{t-1}+\bSigma\big)$ from $\Cb_0=\zero$ gives
\begin{equation}\label{eq:A.5}
\Cb_t=\gamma^2\sum_{r=0}^{t-1}(\Ib-\gamma\Hb)^{t-1-r}\Db_r(\Ib-\gamma\Hb)^{t-1-r},\qquad
\text{where}\quad \Db_r:=(\cM-\wt\cM)\circ\Cb_r+\bSigma.
\end{equation}
Substituting~\eqref{eq:A.5} into~\eqref{eq:A.4} (and taking transposes when $t'<t$), we see that for every pair $(t,t')\in\{s,\ldots,s+N-1\}^2$,
\[
\EE\big[\bmeta_{t'}^{\variance}\otimes\bmeta_t^{\variance}\big]
=\gamma^2\sum_{r=0}^{(t\wedge t')-1}(\Ib-\gamma\Hb)^{t'-1-r}\Db_r(\Ib-\gamma\Hb)^{t-1-r}.
\]
Averaging over all $t,t'\in\{s,\ldots,s+N-1\}$ and exchanging the sum over $r\in\{0,\ldots,s+N-2\}$ with the double sum over $t,t'\ge s\vee(r+1)$ yields
\begin{align*}
\EE\big[\overbar{\bmeta}_{s:s+N}^{\variance}\otimes\overbar{\bmeta}_{s:s+N}^{\variance}\big]
&=\gamma^2\sum_{r=0}^{s+N-2}\left(\frac1N\sum_{t'=s\vee(r+1)}^{s+N-1}(\Ib-\gamma\Hb)^{t'-1-r}\right)\!\Db_r\!\left(\frac1N\sum_{t=s\vee(r+1)}^{s+N-1}(\Ib-\gamma\Hb)^{t-1-r}\right)\\
&=\gamma^2\sum_{r=0}^{s+N-2}\Qb_{r,s,N}\big((\cM-\wt\cM)\circ\Cb_r+\bSigma\big)\Qb_{r,s,N},
\end{align*}
which completes the proof of~\eqref{eq:5.10}.
\end{proof}

\subsection{Two-Sided Spectral Evaluation of Trajectory Functionals}\label{app:A.2}
\begin{lemma}[Two-Sided Spectral Evaluation of $\EffectiveBias$, $\cV_{s,N}$, and $\cS_{s,N}$]\label{lem:A.3}
Suppose Assumption~\ref{ass:3.1} holds and $0<\gamma\le1/(2\lambda_1)$, and recall $\EffectiveBias$, $\cS_{s,N}$, $\cV_{s,N}$, and $(\wb_s^{\drift},\omega_j,\Xi_N)$ from~\eqref{eq:4.1}, \eqref{eq:5.11}, and~\eqref{eq:A.0}. Define the burn-in cross-spectral transfer functional
\begin{equation}\label{eq:A.9}
\Phi_{\burn}(s,N):=\sum_{i,j=1}^{\infty}\min\left\{\frac1{N^2},\gamma^2\lambda_i^2\right\}\lambda_j\omega_j^2\big(1-\gamma(\lambda_i\wedge\lambda_j)\big)^{2(s-1)}\min\left\{s,\frac1{\gamma|\lambda_i-\lambda_j|}\right\}.
\end{equation}
Then for all $s\ge0$ and $N\ge1$ (and $N\ge2$ for the lower bounds on $\cV_{s,N}$ and $\cS_{s,N}$),\footnote{The two-sided spectral evaluation of $\EffectiveBias$ in~\eqref{eq:A.10} already appears in \citet[Lemmas~D.4 and~D.8]{zou2021}; we include its short derivation below to keep our proof self-contained.}
\begin{align}
\frac18\left(\frac1{\gamma^2N^2}\norm{\wb_s^{\drift}}_{\Hb^{-1}_{0:k^*}}^2+\norm{\wb_s^{\drift}}_{\Hb_{k^*:\infty}}^2\right)
&\le\EffectiveBias\notag\\*
&\le\frac12\left(\frac1{\gamma^2N^2}\norm{\wb_s^{\drift}}_{\Hb^{-1}_{0:k^*}}^2+\norm{\wb_s^{\drift}}_{\Hb_{k^*:\infty}}^2\right),\label{eq:A.10}
\end{align}
\begin{equation}\label{eq:A.11}
\frac1{216}\left(\frac{k^*}{N}+\gamma\sum_{k^*<i\le k^\dagger}\lambda_i+(s+N)\gamma^2\sum_{i>k^\dagger}\lambda_i^2\right)
\le\cV_{s,N}\le\frac{k^*}{N}+\gamma\sum_{k^*<i\le k^\dagger}\lambda_i+(s+N)\gamma^2\sum_{i>k^\dagger}\lambda_i^2,
\end{equation}
and
\begin{align}
\cS_{s,N}&\ge\frac1{864}\left(\frac1{\gamma N}\norm{\wb_s^{\drift}}_{\Ib_{0:k^*}}^2+\norm{\wb_s^{\drift}}_{\Hb_{k^*:\infty}}^2\right)\Xi_N+\frac1{108}\Phi_{\burn}(s,N),\notag\\
\cS_{s,N}&\le\frac12\left(\frac1{\gamma N}\norm{\wb_s^{\drift}}_{\Ib_{0:k^*}}^2+\norm{\wb_s^{\drift}}_{\Hb_{k^*:\infty}}^2\right)\Xi_N+\frac12\Phi_{\burn}(s,N).\label{eq:A.12}
\end{align}
\end{lemma}

\begin{proof}[Proof of Lemma~\ref{lem:A.3}]
We establish \eqref{eq:A.10}, \eqref{eq:A.11}, and \eqref{eq:A.12} in three parts.

\subsubsection*{Part 1: Evaluation of $\EffectiveBias$}
For each mode $i\ge1$, the elementary inequalities $\frac12(1\wedge\gamma N\lambda_i)\le(1-e^{-1})(1\wedge\gamma N\lambda_i)\le1-(1-\gamma\lambda_i)^N\le1\wedge(\gamma N\lambda_i)$ imply
\begin{equation}\label{eq:A.13}
\frac14\left(\one_{\{i\le k^*\}}+\gamma^2N^2\lambda_i^2\one_{\{i>k^*\}}\right)
\le\big(1-(1-\gamma\lambda_i)^N\big)^2
\le\one_{\{i\le k^*\}}+\gamma^2N^2\lambda_i^2\one_{\{i>k^*\}}.
\end{equation}
Multiplying~\eqref{eq:A.13} by $\frac{\omega_i^2}{2\gamma^2N^2\lambda_i}(1-\gamma\lambda_i)^{2s}=\frac{(\vb_i^\top\wb_s^{\drift})^2}{2\gamma^2N^2\lambda_i}$ and summing over $i\ge1$ in~\eqref{eq:4.1} directly yields~\eqref{eq:A.10}.

\subsubsection*{Part 2: Evaluation of $\cV_{s,N}$ Across the Three Spectral Regimes}
Using $\Psi_{r,s,N}=\gamma^2F_{r+1}$ from~\eqref{eq:A.0b}, we have $\cV_{s,N}=\frac12\sum_{r=0}^{s+N-2}\Psi_{r,s,N}$. We split this sum into the averaging window $r\in\{s,\ldots,s+N-2\}$ and the burn-in phase $r\in\{0,\ldots,s-1\}$:
\begin{itemize}
\item \emph{Averaging-window contribution ($s\le r\le s+N-2$):} Setting $m=s+N-1-r\in\{1,\ldots,N-1\}$ gives
\begin{align*}
\frac12\sum_{r=s}^{s+N-2}\Psi_{r,s,N}
&=\frac1{2N^2}\sum_{m=1}^{N-1}\sum_{i=1}^{\infty}\bigl(1-(1-\gamma\lambda_i)^m\bigr)^2\\
&\le\frac{N-1}{2N^2}\sum_{i=1}^{\infty}\bigl(1\wedge\gamma^2N^2\lambda_i^2\bigr)
\le\frac12\Xi_N,
\end{align*}
where the first inequality uses $1-(1-\gamma\lambda_i)^m\le1\wedge(\gamma N\lambda_i)$. Conversely, for $N\ge2$, restricting the sum to $r\in\{s,\ldots,s+\floor{N/2}-1\}$ ensures $m\ge\floor{N/2}$, so
\begin{align*}
\frac12\sum_{r=s}^{s+N-2}\Psi_{r,s,N}
&\ge\frac1{2N^2}\sum_{r=s}^{s+\floor{N/2}-1}\sum_{i=1}^{\infty}\bigl(1-(1-\gamma\lambda_i)^{\floor{N/2}}\bigr)^2\\
&\ge\frac{\floor{N/2}}{18N^2}\sum_{i=1}^{\infty}\bigl(1-(1-\gamma\lambda_i)^N\bigr)^2
\ge\frac1{216}\Xi_N,
\end{align*}
where the second inequality applies Lemma~\ref{lem:A.1} and the third inequality uses~\eqref{eq:A.13} together with $\floor{N/2}\ge N/3$ for $N\ge2$. Combining both bounds gives
\begin{equation}\label{eq:A.14}
\frac1{216}\left(\frac{k^*}{N}+N\gamma^2\sum_{i>k^*}\lambda_i^2\right)
\le\frac12\sum_{r=s}^{s+N-2}\Psi_{r,s,N}
\le\frac12\left(\frac{k^*}{N}+N\gamma^2\sum_{i>k^*}\lambda_i^2\right).
\end{equation}

\item \emph{Burn-in contribution ($0\le r\le s-1$, when $s\ge1$):} Summing the geometric series over $0\le r\le s-1$ yields
\[
\frac12\sum_{r=0}^{s-1}\Psi_{r,s,N}
=\frac1{2N^2}\sum_{i=1}^{\infty}\bigl(1-(1-\gamma\lambda_i)^N\bigr)^2\frac{1-(1-\gamma\lambda_i)^{2s}}{\gamma\lambda_i(2-\gamma\lambda_i)}.
\]
Since $\frac14(s\wedge(\gamma\lambda_i)^{-1})\le\frac{1-(1-\gamma\lambda_i)^{2s}}{\gamma\lambda_i(2-\gamma\lambda_i)}\le s\wedge(\gamma\lambda_i)^{-1}$, applying~\eqref{eq:A.13} shows that $\frac12\sum_{r=0}^{s-1}\Psi_{r,s,N}$ is sandwiched between $\frac1{32}\sum_{i\ge1}v_i$ and $\frac12\sum_{i\ge1}v_i$, where $v_i:=\frac1{N^2}(1\wedge\gamma^2N^2\lambda_i^2)\min\{s,(\gamma\lambda_i)^{-1}\}$. Adapting the three-regime spectral partitioning strategy of \citet[Lemma~D.3]{zou2021}, we evaluate $v_i$ across the cutoffs $k^*$ and $k^\dagger$:
\begin{enumerate}
\item \emph{Case 1 ($i\le k^*$, so $\gamma\lambda_i\ge N^{-1}$):} Here $1\wedge\gamma^2N^2\lambda_i^2=1$ and $\min\{s,(\gamma\lambda_i)^{-1}\}\le(\gamma\lambda_i)^{-1}\le N$, so $0\le v_i\le\frac1N$.
\item \emph{Case 2 ($k^*<i\le k^\dagger$, so $(s+N)^{-1}\le\gamma\lambda_i<N^{-1}$):} Here $1\wedge\gamma^2N^2\lambda_i^2=\gamma^2N^2\lambda_i^2$, so $v_i=\gamma^2\lambda_i^2\min\{s,(\gamma\lambda_i)^{-1}\}$. In the upper bound, $\min\{s,(\gamma\lambda_i)^{-1}\}\le(\gamma\lambda_i)^{-1}$ gives $v_i\le\gamma\lambda_i$. In the lower bound, if $s\ge(\gamma\lambda_i)^{-1}$, then $v_i=\gamma\lambda_i$; if $s<(\gamma\lambda_i)^{-1}\le s+N$, then adding the averaging-window contribution $N\gamma^2\lambda_i^2$ from~\eqref{eq:A.14} gives $v_i+N\gamma^2\lambda_i^2=(s+N)\gamma^2\lambda_i^2\ge\gamma\lambda_i$.
\item \emph{Case 3 ($i>k^\dagger$, so $\gamma\lambda_i<(s+N)^{-1}\le s^{-1}$):} Here $1\wedge\gamma^2N^2\lambda_i^2=\gamma^2N^2\lambda_i^2$ and $\min\{s,(\gamma\lambda_i)^{-1}\}=s$, so $v_i=s\gamma^2\lambda_i^2$.
\end{enumerate}
\end{itemize}
Summing Cases 1--3 and combining with~\eqref{eq:A.14} (noting that for $k^*<i\le k^\dagger$, $N\gamma^2\lambda_i^2\le\gamma\lambda_i$ in the upper bound) yields~\eqref{eq:A.11}.

\subsubsection*{Part 3: Evaluation of $\cS_{s,N}$}
Writing $b_r=\norm{(\Ib-\gamma\Hb)^r(\wb_0-\wb^*)}_{\Hb}^2=\sum_{j\ge1}\lambda_j\omega_j^2(1-\gamma\lambda_j)^{2r}$, we have $\cS_{s,N}=\frac12\sum_{r=0}^{s+N-2}b_r\Psi_{r,s,N}$. We split this sum into the averaging window $r=s+k$ ($0\le k\le N-2$) and the burn-in phase $0\le r\le s-1$:
\begin{itemize}
\item \emph{Averaging-window contribution ($r=s+k$, $0\le k\le N-2$):} For $N\ge2$,\footnote{When $N=1$, this sum is empty and its upper bound in~\eqref{eq:A.15} holds trivially.} note first from Part~2 that $\Psi_{s+k,s,N}\le\Psi_{s,s,N}\le\frac1N\Xi_N$ for all $0\le k\le N-2$, while Lemma~\ref{lem:A.1} and~\eqref{eq:A.13} give $\Psi_{s+k,s,N}\ge\frac1{36N}\Xi_N$ for all $0\le k\le\floor{N/2}-1$. Moreover, since $b_{s+k}=\sum_{j\ge1}\lambda_j(\vb_j^\top\wb_s^{\drift})^2(1-\gamma\lambda_j)^{2k}$, summing the geometric series over $k$ gives
\begin{align*}
\gamma\sum_{k=0}^{N-1}b_{s+k}
&=\sum_{j=1}^{\infty}\frac{1-(1-\gamma\lambda_j)^{2N}}{2-\gamma\lambda_j}(\vb_j^\top\wb_s^{\drift})^2\\
&\le\norm{\wb_s^{\drift}}_{\Ib_{0:k^*}}^2+\gamma N\norm{\wb_s^{\drift}}_{\Hb_{k^*:\infty}}^2,
\end{align*}
where the inequality uses $\frac{1-(1-\gamma\lambda_j)^{2N}}{2-\gamma\lambda_j}\le1$ on $j\le k^*$ (since $2-\gamma\lambda_j\ge1$) and $\gamma\sum_{k=0}^{N-1}\lambda_j(1-\gamma\lambda_j)^{2k}\le\gamma N\lambda_j$ on $j>k^*$. Conversely, restricting the sum to $0\le k\le\floor{N/2}-1$ yields
\begin{align*}
\gamma\sum_{k=0}^{\floor{N/2}-1}b_{s+k}
&=\sum_{j=1}^{\infty}\frac{1-(1-\gamma\lambda_j)^{2\floor{N/2}}}{2-\gamma\lambda_j}(\vb_j^\top\wb_s^{\drift})^2\\
&\ge\frac16\sum_{j=1}^{\infty}\big[1-(1-\gamma\lambda_j)^{2N}\big](\vb_j^\top\wb_s^{\drift})^2\\
&\ge\frac1{12}\left(\norm{\wb_s^{\drift}}_{\Ib_{0:k^*}}^2+\gamma N\norm{\wb_s^{\drift}}_{\Hb_{k^*:\infty}}^2\right),
\end{align*}
where the first inequality applies Lemma~\ref{lem:A.1} together with $2-\gamma\lambda_j\le2$, and the second inequality uses $1-(1-\gamma\lambda_j)^{2N}\ge1-(1-\gamma\lambda_j)^N\ge\frac12(1\wedge\gamma N\lambda_j)$. Combining the filter and geometric-sum bounds yields
\begin{align}
\frac1{864}\left(\frac1{\gamma N}\norm{\wb_s^{\drift}}_{\Ib_{0:k^*}}^2+\norm{\wb_s^{\drift}}_{\Hb_{k^*:\infty}}^2\right)\Xi_N
&\le\frac12\sum_{k=0}^{N-2}b_{s+k}\Psi_{s+k,s,N}\notag\\
&\le\frac12\left(\frac1{\gamma N}\norm{\wb_s^{\drift}}_{\Ib_{0:k^*}}^2+\norm{\wb_s^{\drift}}_{\Hb_{k^*:\infty}}^2\right)\Xi_N.\label{eq:A.15}
\end{align}

\item \emph{Burn-in contribution ($0\le r\le s-1$, when $s\ge1$):} Expanding $b_r$ and $\Psi_{r,s,N}$ gives
\[
\frac12\sum_{r=0}^{s-1}b_r\Psi_{r,s,N}
=\frac1{2N^2}\sum_{i=1}^{\infty}\big(1-(1-\gamma\lambda_i)^N\big)^2\sum_{j=1}^{\infty}\lambda_j\omega_j^2\sum_{r=0}^{s-1}(1-\gamma\lambda_j)^{2r}(1-\gamma\lambda_i)^{2(s-1-r)}.
\]
For each mode pair $(i,j)$, let $p:=(1-\gamma\lambda_j)^2$ and $q:=(1-\gamma\lambda_i)^2$, so that $M:=p\vee q=(1-\gamma(\lambda_i\wedge\lambda_j))^2\in[1/4,1]$ (since $\gamma\lambda_1\le1/2$) and $M-m=\gamma|\lambda_i-\lambda_j|(2-\gamma(\lambda_i+\lambda_j))\in[\gamma|\lambda_i-\lambda_j|,2\gamma|\lambda_i-\lambda_j|]$. Consequently, $\frac1{8\gamma|\lambda_i-\lambda_j|}\le\frac{M}{M-m}\le\frac1{\gamma|\lambda_i-\lambda_j|}$. Applying Lemma~\ref{lem:A.2} to the inner geometric convolution over $r$ and~\eqref{eq:A.13} to $(1-(1-\gamma\lambda_i)^N)^2$ term-by-term yields
\begin{equation}\label{eq:A.16}
\frac1{108}\Phi_{\burn}(s,N)
\le\frac{1-e^{-1}}{64}\Phi_{\burn}(s,N)
\le\frac12\sum_{r=0}^{s-1}b_r\Psi_{r,s,N}
\le\frac12\Phi_{\burn}(s,N).
\end{equation}
\end{itemize}
Adding~\eqref{eq:A.15} and~\eqref{eq:A.16} establishes~\eqref{eq:A.12}.
\end{proof}

\section{Proofs of Main Theorems, Propositions, and Corollaries}\label{app:B}

\subsection{Proof of Theorem \ref{thm:4.1}, Theorem \ref{thm:4.2}, and Corollary \ref{cor:4.5}}\label{app:B.1}

\begin{proof}[Proof of Theorem~\ref{thm:4.1} and Theorem~\ref{thm:4.2}]
For Theorem~\ref{thm:4.1}, Lemma~\ref{lem:C.3}(1) gives $\EE[L(\overbar{\wb}_{s:s+N})]-L(\wb^*)\allowbreak\le 2\cdot\bias+2\cdot\variance=2\cdot\EffectiveBias+2\cdot\EffectiveVar$ under Assumptions~\ref{ass:3.1}--\ref{ass:3.3} (and equality without the factor of $2$ under the well-specified model \eqref{eq:3.4} by Lemma~\ref{lem:C.3}(2)). Invoking Lemma~\ref{lem:5.2}(1) and substituting the upper bounds \eqref{eq:A.10}--\eqref{eq:A.12} from Lemma~\ref{lem:A.3} directly establishes \eqref{eq:4.4} and \eqref{eq:4.5}.

For Theorem~\ref{thm:4.2}, Lemma~\ref{lem:C.3}(2) gives the exact decomposition $\EE[L(\overbar{\wb}_{s:s+N})]-L(\wb^*)=\EffectiveBias+\EffectiveVar$ under \eqref{eq:3.4}. Invoking Lemma~\ref{lem:5.2}(2) and substituting the lower bounds \eqref{eq:A.10}--\eqref{eq:A.12} from Lemma~\ref{lem:A.3} directly establishes \eqref{eq:4.7} and \eqref{eq:4.8}.
\end{proof}

\begin{proof}[Proof of Corollary~\ref{cor:4.5}]
Suppose $c_0N\le s\le C_0N$, and recall $\wb_s^{\drift}$, $\omega_j$, and $\Phi_{\burn}(s,N)$ from \eqref{eq:A.0} and \eqref{eq:A.9}. For every slow mode $i>k^*$, we have $\gamma\lambda_i<N^{-1}\le C_0/s$, which implies $(1-\gamma\lambda_i)^{2s}\in[e^{-4C_0},1]$ and hence $\norm{\wb_s^{\drift}}_{\Hb_{k^*:\infty}}^2\asymp\norm{\wb_0-\wb^*}_{\Hb_{k^*:\infty}}^2$. Similarly, for every intermediate mode $i\in(k^*,k^\dagger]$, we have $((1+C_0)N)^{-1}\le\gamma\lambda_i<N^{-1}$, so $\gamma\lambda_i\asymp N\gamma^2\lambda_i^2\asymp N^{-1}$.

To evaluate $\Phi_{\burn}(s,N)$ in \eqref{eq:A.9}, we partition the double sum over $(i,j)$ into three blocks:
\begin{itemize}
\item \emph{Fast-to-fast block ($i\le k^*$ and $j\le k^*$):} Since $\min\{N^{-2},\gamma^2\lambda_i^2\}=N^{-2}$ on $i\le k^*$, this block equals the third line of \eqref{eq:4.16}.
\item \emph{Slow-to-fast block ($i\le k^*$ and $j>k^*$):} Using $\min\{N^{-2},\gamma^2\lambda_i^2\}=N^{-2}$ and $(1-\gamma(\lambda_i\wedge\lambda_j))^{2(s-1)}\le1$, we upper-bound this block by
\begin{align*}
\frac1{N^2}\sum_{i\le k^*}\sum_{j>k^*}\lambda_j\omega_j^2\min\left\{s,\frac1{\gamma|\lambda_i-\lambda_j|}\right\}
&\le\frac{2\vee C_0}{N}\sum_{i\le k^*}\sum_{j>k^*}\lambda_j\omega_j^2\\
&=(2\vee C_0)\norm{\wb_0-\wb^*}_{\Hb_{k^*:\infty}}^2\frac{k^*}{N},
\end{align*}
where the inequality uses $\min\{s,(\gamma|\lambda_i-\lambda_j|)^{-1}\}\le2(\gamma\lambda_i)^{-1}\le2N$ when $\lambda_j\le\lambda_i/2$ and $\min\{s,(\gamma|\lambda_i-\lambda_j|)^{-1}\}\le s\le C_0N$ when $\lambda_j>\lambda_i/2$. This contribution is absorbed into the second line of \eqref{eq:4.16}.
\item \emph{Slow receiving block ($i>k^*$ and $j\ge1$):} On $i>k^*$, we have $\min\{N^{-2},\gamma^2\lambda_i^2\}=\gamma^2\lambda_i^2$ and $\lambda_i\wedge\lambda_j\le\lambda_i<(\gamma N)^{-1}$, so $(1-\gamma(\lambda_i\wedge\lambda_j))^{2(s-1)}\in[e^{-4C_0},1]$. Moreover, for $j\le k^*$, we have $\min\{s,(\gamma|\lambda_i-\lambda_j|)^{-1}\}\asymp(\gamma\lambda_j)^{-1}$ (both when $\lambda_j\ge2\lambda_i$ and when $\lambda_i<\lambda_j<2\lambda_i$, since $s\asymp N\asymp(\gamma\lambda_j)^{-1}$ in the latter case), whereas for $j>k^*$, both $\gamma\lambda_i<N^{-1}$ and $\gamma\lambda_j<N^{-1}$ imply $\min\{s,(\gamma|\lambda_i-\lambda_j|)^{-1}\}\asymp N$. Summing over $j\ge1$ and $i>k^*$ therefore gives
\begin{align*}
&\sum_{i>k^*}\gamma^2\lambda_i^2\sum_{j=1}^{\infty}\lambda_j\omega_j^2\bigl(1-\gamma(\lambda_i\wedge\lambda_j)\bigr)^{2(s-1)}\min\left\{s,\frac1{\gamma|\lambda_i-\lambda_j|}\right\}\\
&\quad\asymp\left(\sum_{i>k^*}\gamma^2\lambda_i^2\right)\left(\frac1\gamma\norm{\wb_0-\wb^*}_{\Ib_{0:k^*}}^2+N\norm{\wb_0-\wb^*}_{\Hb_{k^*:\infty}}^2\right)\\
&\quad=\frac{\norm{\wb_0-\wb^*}_{\Ib_{0:k^*}}^2+N\gamma\norm{\wb_0-\wb^*}_{\Hb_{k^*:\infty}}^2}{N\gamma}\cdot\left(N\gamma^2\sum_{i>k^*}\lambda_i^2\right),
\end{align*}
which yields the first line of \eqref{eq:4.16}.
\end{itemize}
Substituting these evaluations into Theorems~\ref{thm:4.1} and~\ref{thm:4.2} establishes \eqref{eq:4.15}--\eqref{eq:4.16}.
\end{proof}

\subsection{Proof of Proposition \ref{prop:4.3} and Corollary \ref{cor:4.4}}\label{app:B.2}

\begin{proof}[Proof of Proposition~\ref{prop:4.3}]
Expanding $q_t(\lambda_i)^2=\frac1{N^2}\sum_{a=s\vee t}^T\sum_{b=s\vee t}^T(1-\gamma\lambda_i)^{a+b-2t}$ in \eqref{eq:4.2} and summing against $\lambda_i^2$ yields the pair-midpoint identity \eqref{eq:4.10}. When $T\le1$, the maximum in \eqref{eq:4.3} is empty so $\chi_{s,N}=0$. Suppose $T\ge2$ and fix any $r\in\{1,\ldots,T-1\}$. Since $s\vee t\ge s\vee(r+1)$ for all $t\ge r+1$, substituting \eqref{eq:4.10} into $\gamma^2\sum_{t=r+1}^Tk(t-r-1)F_t$ and exchanging the order of summation gives
\begin{equation}\label{eq:B.0}
\gamma^2\sum_{t=r+1}^Tk(t-r-1)F_t
=\frac{\gamma^2}{N^2}\sum_{a=s\vee(r+1)}^T\sum_{b=s\vee(r+1)}^T\sum_{t=r+1}^{\min(a,b)}k(t-r-1)k\Bigl(\frac{a+b}{2}-t\Bigr).
\end{equation}
For each pair $(a,b)$, set $j:=t-r-1\ge0$ and $h:=\frac{a+b}{2}-r\in[1,T]$, so the inner sum in \eqref{eq:B.0} runs over $0\le j\le\min(a,b)-r-1\le T-2$ with $h-1-j=\frac{a+b}{2}-t\ge0$.

For part (1), for every $0\le j\le\min(a,b)-r-1$, we have
\begin{align*}
k(j)k(h-1-j)
&\le\tr(\Hb^2)k(h-1)\\
&=\tr(\Hb^2)\sum_{i=1}^{\infty}\lambda_i^2(1-\gamma\lambda_i)^{2h-2}
\le\frac{\tr(\Hb^2)}{(1-\gamma\lambda_1)^2}k(h),
\end{align*}
where the first inequality follows from Chebyshev's sum inequality (Lemma~\ref{lem:C.0} with $u=j$ and $v=h-1-j$) and the second inequality uses $(1-\gamma\lambda_i)^{-2}\le(1-\gamma\lambda_1)^{-2}$. Since the inner sum over $j$ in \eqref{eq:B.0} contains at most $T-1$ terms and $\frac1{N^2}\sum_{a,b=s\vee(r+1)}^Tk(\frac{a+b}{2}-r)\le\frac1{N^2}\sum_{a,b=s\vee r}^Tk(\frac{a+b}{2}-r)=F_r$, summing over $a,b$ yields $\gamma^2\sum_{t=r+1}^Tk(t-r-1)F_t\le\frac{\gamma^2(T-1)\tr(\Hb^2)}{(1-\gamma\lambda_1)^2}F_r$, which establishes \eqref{eq:4.11} and \eqref{eq:4.12}.

For part (2), since $k(\cdot)$ is non-negative and non-increasing on $[0,\infty)$, splitting the inner convolution over $0\le j\le\min(a,b)-r-1\le h-1$ at the midpoint $(h-1)/2$ yields
\begin{align*}
\sum_{j=0}^{\min(a,b)-r-1}k(j)k(h-1-j)
&\le\sum_{0\le j\le(h-1)/2}k(j)k\Bigl(\frac{h-1}{2}\Bigr)
+\sum_{(h-1)/2<j\le h-1}k\Bigl(\frac{h-1}{2}\Bigr)k(h-1-j)\\
&\le2k\Bigl(\frac{h-1}{2}\Bigr)\sum_{n=0}^\infty k(n)
\le2D_Tk(h)\sum_{n=0}^\infty k(n),
\end{align*}
where the first inequality bounds $k(h-1-j)\le k((h-1)/2)$ on the first half and $k(j)\le k((h-1)/2)$ on the second half, and the third inequality uses $k((h-1)/2)\le D_Tk(h)$ from \eqref{eq:4.9}. Substituting this bound into \eqref{eq:B.0} and using $\frac1{N^2}\sum_{a,b=s\vee(r+1)}^Tk(\frac{a+b}{2}-r)\le F_r$ gives
\[
\gamma^2\sum_{t=r+1}^Tk(t-r-1)F_t\le2D_T\gamma^2\biggl(\sum_{n=0}^\infty k(n)\biggr)F_r.
\]
Multiplying by $\alpha/F_r$, taking the maximum over $1\le r<T$, and evaluating the geometric series $\gamma^2\sum_{n=0}^\infty k(n)=\gamma\sum_{i\ge1}\frac{\lambda_i}{2-\gamma\lambda_i}\le\frac{\gamma\tr(\Hb)}{2-\gamma\lambda_1}$ establishes \eqref{eq:4.13}.
\end{proof}

\begin{proof}[Proof of Corollary~\ref{cor:4.4}]
Let $\nu:=2-1/a\in(1,2)$ and $z:=\gamma\Lambda h\ge0$. Since $\gamma\lambda_i\le\gamma\Lambda\le1/2$, the elementary inequalities $-2x\le\log(1-x)\le-x$ for $x\in[0,1/2]$ imply $e^{-4\gamma h\lambda_i}\le(1-\gamma\lambda_i)^{2h}\le e^{-2\gamma h\lambda_i}$.
\begin{itemize}
\item \emph{Upper bound on $k(h)$:} For $z\in[0,1]$, $k(h)\le k(0)\le\Lambda^2\sum_{i\ge1}i^{-2a}\le2^\nu\Lambda^2(\sum_{i\ge1}i^{-2a})(1+z)^{-\nu}$. For $z>1$, set $M:=\ceil{z^{1/a}}\ge2$ and split the sum at $M$:
\begin{align*}
k(h)&\le\Lambda^2\sum_{i=1}^{M}i^{-2a}e^{-2\theta zi^{-a}}+\Lambda^2\sum_{i=M+1}^{\infty}i^{-2a}\\
&\le\Lambda^2(\theta z)^{-2}\sum_{i=1}^{M}1+\Lambda^2\int_{M}^{\infty}x^{-2a}\ud x\\
&\le\Lambda^2\left(2\theta^{-2}+\frac1{2a-1}\right)z^{-\nu}
\le C(a,\theta)\Lambda^2(1+\gamma\Lambda h)^{-\nu},
\end{align*}
where the second inequality uses $v^2e^{-v}\le4e^{-2}\le1$ with $v=\theta zi^{-a}$ on $1\le i\le M$ and integral comparison on $i>M$, the third inequality uses $M=\ceil{z^{1/a}}\le2z^{1/a}$ and $M^{1-2a}\le z^{-\nu}$, and the last inequality uses $z^{-\nu}\le2^\nu(1+z)^{-\nu}$ for $z>1$.
\item \emph{Lower bound on $k(h)$:} For $z\in[0,1]$, retaining $i=1$ gives $k(h)\ge\theta^2\Lambda^2e^{-4z}\ge\theta^2e^{-4}\Lambda^2(1+z)^{-\nu}$. For $1\le z\le(d/2)^a$ (with $d=\infty$ in infinite dimension), let $m:=\floor{z^{1/a}}\ge1$, so that $2m\le d$ and $m+1>z^{1/a}$. Restricting the sum to $m+1\le i\le2m$, where $zi^{-a}\le z(m+1)^{-a}<1$, gives
\begin{align*}
k(h)&\ge\theta^2\Lambda^2\sum_{i=m+1}^{2m}i^{-2a}e^{-4zi^{-a}}
\ge\theta^2\Lambda^2e^{-4}\sum_{i=m+1}^{2m}i^{-2a}\\
&\ge\theta^2\Lambda^2e^{-4}m(2m)^{-2a}
\ge\theta^2e^{-4}2^{-2a}\Lambda^2(1+\gamma\Lambda h)^{-\nu},
\end{align*}
where the last inequality uses $m\le z^{1/a}\le(1+z)^{1/a}$ and $1-2a<0$.
\end{itemize}
Combining the upper and lower kernel bounds, for every $1\le h\le T$ (with $\gamma\Lambda T\le(d/2)^a$),
\[
\frac{k((h-1)/2)}{k(h)}\le\frac{C(a,\theta)}{\theta^2e^{-4}2^{-2a}}
\biggl(\frac{1+\gamma\Lambda h}{1+\gamma\Lambda(h-1)/2}\biggr)^\nu
\le\frac{2^\nu C(a,\theta)}{\theta^2e^{-4}2^{-2a}}=:D_*(a,\theta).
\]
Consequently, $D_T\le D_*$, and for any $\gamma$ satisfying \eqref{eq:4.14}, Proposition~\ref{prop:4.3}(2) yields $\chi_{s,N}\le\frac{2\alpha D_*\gamma\tr(\Hb)}{2-\gamma\lambda_1}\le2\alpha D_*\gamma\tr(\Hb)\le1/2$.
\end{proof}

\subsection{Proof of Proposition \ref{prop:4.7} and Proposition \ref{prop:4.8}}\label{app:B.3}

\begin{proof}[Proof of Proposition~\ref{prop:4.7}]
For part (1), we have $\lambda_1=1$, $\lambda_2=(4\gamma N)^{-1}$, $\omega_1=1$, $\omega_2=0$, $\sigma_{\noise}^2=0$, and $\beta=1$ (since $\xb\sim\cN(\zero,\Hb)$). Let $m:=\floor{1/(4\gamma)}+1\le N$. For every $1\le t\le m$, we have $t-1\le1/(4\gamma)$, so Bernoulli's inequality gives $(1-\gamma)^{2(t-1)}\ge1-2\gamma(t-1)\ge1/2$. Meanwhile, for every $1\le t\le N$ and $N\le k\le2N-1$, we have $k-t\le2N$, so $(1-\gamma\lambda_2)^{k-t}=(1-\frac1{4N})^{k-t}\ge1-\frac{k-t}{4N}\ge1/2$, which implies $q_t(\lambda_2)=\frac1N\sum_{k=N}^{2N-1}(1-\gamma\lambda_2)^{k-t}\ge1/2$ and $F_t\ge\lambda_2^2q_t(\lambda_2)^2\ge\frac14(4\gamma N)^{-2}=\frac1{64\gamma^2N^2}$.
Substituting these lower bounds into \eqref{eq:5.13} of Lemma~\ref{lem:5.2} with $\beta=1$ yields
\begin{align*}
\EE\bigl[L(\overbar{\wb}_{N:2N})\bigr]-L(\wb^*)
&\ge\cS_{N,N}
\ge\frac{\gamma^2}{2}\sum_{t=1}^m(1-\gamma)^{2(t-1)}F_t\\
&\ge\frac{\gamma^2}{2}\cdot m\cdot\frac12\cdot\frac1{64\gamma^2N^2}
\ge\frac1{1024\gamma N^2},
\end{align*}
where the last inequality uses $m>1/(4\gamma)$. By contrast, since $\lambda_1=1\ge(2\gamma N)^{-1}>\lambda_2=(4\gamma N)^{-1}$, we have $k^*=k^\dagger=1$ and $\norm{\wb_0-\wb^*}_{\Hb_{k^\dagger:\infty}}^2=\lambda_2\omega_2^2=0$, so \citet[Theorem~5.2]{zou2021} drops the variance contribution entirely and retains only the bias lower bound $\frac1{100\gamma^2N^2}\norm{(\Ib-\gamma\Hb)^N(\wb_0-\wb^*)}_{\Hb_{0:k^*}^{-1}}^2=\frac{(1-\gamma)^{2N}}{100\gamma^2N^2}$, proving \eqref{eq:4.17}.

For part (2), both $\cD_1$ and $\cD_2$ satisfy $\EE[x^2]=1$, $\EE[x^4]\le3=\alpha$, and $\EE[(x^2-1)^2]=\kappa-1\ge1=\beta$ for $\kappa:=\EE[x^4]\in\{2,3\}$. Since $y_t=0$ and $w_0=1$, the scalar iterate satisfies $w_t=(1-\gamma x_t^2)w_{t-1}$ with $\EE[1-\gamma x_t^2]=1-\gamma>0$ and $\EE[(1-\gamma x_t^2)^2]=1-2\gamma+\kappa\gamma^2=r_\kappa\in(0,1)$ for $0<\gamma<1/3$. By independence of $\{x_t\}$, we have $\EE[w_t^2]=r_\kappa^t$ and $\EE[w_tw_{t+h}]=r_\kappa^t(1-\gamma)^h$ for all $t,h\ge0$. Expanding $\overbar w_{s:s+N}^2=\frac1{N^2}\sum_{t=s}^{s+N-1}w_t^2+\frac2{N^2}\sum_{t=s}^{s+N-1}\sum_{h=1}^{s+N-1-t}w_tw_{t+h}$ gives the exact risk
\[
R_\kappa(s,N)=\frac12\EE\bigl[\overbar w_{s:s+N}^2\bigr]
=\frac{r_\kappa^s}{2N^2}\sum_{j=0}^{N-1}r_\kappa^j
\left(1+2\sum_{h=1}^{N-1-j}(1-\gamma)^h\right).
\]
Retaining $j=0$ and dropping $h\ge1$ gives $R_\kappa(s,N)\ge\frac{r_\kappa^s}{2N^2}$, while bounding $1+2\sum_{h=1}^\infty(1-\gamma)^h=\frac{2-\gamma}{\gamma}$ and $\sum_{j=0}^\infty r_\kappa^j=\frac1{1-r_\kappa}$ gives $R_\kappa(s,N)\le\frac{(2-\gamma)r_\kappa^s}{2\gamma(1-r_\kappa)N^2}$, establishing \eqref{eq:4.18}.
\end{proof}

\begin{proof}[Proof of Proposition~\ref{prop:4.8}]
Applying the martingale-difference expansion of Lemma~\ref{lem:5.1} to the shifted bias process $\{\bmeta_{s+k}^{\bias}\}_{k=0}^{N-1}$ initialized at $\bmeta_s^{\bias}$ with $\EE[\bmeta_s^{\bias}\otimes\bmeta_s^{\bias}]=\Bb_s$ yields the conditional identity
\begin{equation}\label{eq:B.1}
\EE\bigl[\overbar{\bmeta}_{s:s+N}^{\bias}\otimes\overbar{\bmeta}_{s:s+N}^{\bias}\bigr]
=\Pb_{0,N}\Bb_s\Pb_{0,N}
+\gamma^2\sum_{k=0}^{N-2}\Qb_{k,0,N}\bigl((\cM-\wt\cM)\circ\Bb_{s+k}\bigr)\Qb_{k,0,N}.
\end{equation}
Let $a_i^{(s)}:=\vb_i^\top\Bb_s\vb_i\ge0$ denote the $i$-th diagonal entry of $\Bb_s$ in the eigenbasis of $\Hb$, and recall $\Xi_N$ and $\Psi_{r,s,N}$ from \eqref{eq:A.0}--\eqref{eq:A.0b}. Taking the trace inner product of \eqref{eq:B.1} with $\frac12\Hb$ gives
\[
\bias=\frac12\la\Hb,\Pb_{0,N}\Bb_s\Pb_{0,N}\ra
+\frac{\gamma^2}{2}\sum_{k=0}^{N-2}\inner{\Hb}{\Qb_{k,0,N}\bigl((\cM-\wt\cM)\circ\Bb_{s+k}\bigr)\Qb_{k,0,N}}.
\]
For the first term, since $\Pb_{0,N}\vb_i=\frac{1-(1-\gamma\lambda_i)^N}{\gamma N\lambda_i}\vb_i$, applying \eqref{eq:A.13} yields
\begin{equation}\label{eq:B.2}
\begin{aligned}
\frac1{8\gamma^2N^2}\tr\bigl(\Hb_{0:k^*}^{-1}\Bb_s\bigr)+\frac18\tr\bigl(\Hb_{k^*:\infty}\Bb_s\bigr)
&\le\frac12\la\Hb,\Pb_{0,N}\Bb_s\Pb_{0,N}\ra\\
&\le\frac1{2\gamma^2N^2}\tr\bigl(\Hb_{0:k^*}^{-1}\Bb_s\bigr)+\frac12\tr\bigl(\Hb_{k^*:\infty}\Bb_s\bigr).
\end{aligned}
\end{equation}
For the second term, note that $\gamma^2\la\Hb,\Qb_{k,0,N}\Hb\Qb_{k,0,N}\ra\allowbreak=\Psi_{s+k,s,N}$.\footnote{For $N=1$, the averaging-window innovation sum is empty. The following window calculation treats $N\ge2$; the variance bound applies for every $N\ge1$.} In the upper bound, we have
\begin{align*}
&\frac{\gamma^2}{2}\sum_{k=0}^{N-2}\inner{\Hb}{\Qb_{k,0,N}\bigl((\cM-\wt\cM)\circ\Bb_{s+k}\bigr)\Qb_{k,0,N}}\\
&\quad\le\frac\alpha2\sum_{k=0}^{N-2}\tr(\Hb\Bb_{s+k})\Psi_{s+k,s,N}
\le\frac{\alpha\Xi_N}{4\gamma N(1-\gamma\alpha\tr(\Hb))}\tr(\Bb_{s,s+N})\\
&\quad\le\frac\alpha{2(1-\gamma\alpha\tr(\Hb))}
\left(\frac1{\gamma N}\tr\bigl(\Ib_{0:k^*}\Bb_s\bigr)+\tr\bigl(\Hb_{k^*:\infty}\Bb_s\bigr)\right)\Xi_N,
\end{align*}
where the first inequality applies Assumption~\ref{ass:3.2}, the second inequality uses $\Psi_{s+k,s,N}\le\Psi_{s,s,N}\le\frac1N\Xi_N$ and Lemma~\ref{lem:C.4} on $[s,s+N]$, and the third inequality uses $1-(1-\gamma\lambda_i)^{2N}\le1\wedge(2\gamma N\lambda_i)$. Meanwhile, for the variance error in \eqref{eq:5.10}, Lemma~\ref{lem:C.5} and Assumption~\ref{ass:3.2} imply $(\cM-\wt\cM)\circ\Cb_r+\bSigma\preceq\frac{\sigma^2}{1-\gamma\alpha\tr(\Hb)}\Hb$ for all $r\ge0$, so $\variance\le\frac{\sigma^2}{1-\gamma\alpha\tr(\Hb)}\cV_{s,N}$. Combining these bounds with $\EE[L(\overbar{\wb}_{s:s+N})]-L(\wb^*)\le2\cdot\bias+2\cdot\variance$ and \eqref{eq:A.11} yields \eqref{eq:4.19}.

In the lower bound ($N\ge2$ and $\gamma\le1/(2\lambda_1)$), let $m:=\floor{N/2}\ge1$ and $\Gb:=\Ib-\gamma\Hb$. Restricting the sum over $k$ to $0\le k\le m-1$, we have $\Psi_{s+k,s,N}\ge\frac1{36N}\Xi_N$ from Part~3 of the proof of Lemma~\ref{lem:A.3}, while Assumption~\ref{ass:3.4} gives $(\cM-\wt\cM)\circ\Bb_{s+k}\succeq\beta\tr(\Hb\Bb_{s+k})\Hb$. Summing $\gamma\tr(\Hb\Bb_{s+k})$ over $0\le k\le m-1$ gives
\begin{align*}
\gamma\sum_{k=0}^{m-1}\tr(\Hb\Bb_{s+k})
&\ge\gamma\sum_{k=0}^{m-1}\tr(\Hb\Gb^{2k}\Bb_s)
=\sum_{i=1}^\infty\frac{1-(1-\gamma\lambda_i)^{2m}}{2-\gamma\lambda_i}a_i^{(s)}\\
&\ge\frac12\tr(\Bb_s-\Gb^{2m}\Bb_s)\ge\frac16\tr(\Bb_{s,s+N})\\
&\ge\frac1{12}\Bigl(\tr\bigl(\Ib_{0:k^*}\Bb_s\bigr)+N\gamma\tr\bigl(\Hb_{k^*:\infty}\Bb_s\bigr)\Bigr),
\end{align*}
where the first inequality uses $\Bb_{s+k}=(\cI-\gamma\cT)^k\circ\Bb_s\succeq\Gb^k\Bb_s\Gb^k$ from Lemma~\ref{lem:C.2}(3), the second inequality uses $2-\gamma\lambda_i\le2$, the third inequality applies Lemma~\ref{lem:A.1}, and the fourth inequality uses $1-(1-\gamma\lambda_i)^{2N}\ge\frac12(1\wedge\gamma N\lambda_i)$. Multiplying this trace bound by $\frac\beta{2\gamma}\cdot\frac1{36N}\Xi_N=\frac\beta{72\gamma N}\Xi_N$ gives $\frac\beta{864\gamma N}\bigl(\tr(\Ib_{0:k^*}\Bb_s)+N\gamma\tr(\Hb_{k^*:\infty}\Bb_s)\bigr)\Xi_N$. Combining this with \eqref{eq:B.2}, $\variance\ge\sigma_{\noise}^2\cV_{s,N}$, and \eqref{eq:A.11} establishes \eqref{eq:4.20}.
\end{proof}

\section{Auxiliary Lemmas}\label{app:C}

In this appendix, we collect the general-purpose elementary algebraic and convolution lemmas (Section~\ref{app:C.1}) and the foundational operator lemmas from \citet{jain2017a,jain2017b} and \citet{zou2021} (Section~\ref{app:C.2}) invoked in Appendices~\ref{app:A}--\ref{app:B}.

\subsection{Elementary Algebraic and Convolution Lemmas}\label{app:C.1}

\begin{lemma}[Chebyshev's Sum Inequality for Weighted Monotone Sequences]\label{lem:C.0}
Let $\{w_i\}_{i\ge1}$ be non-negative weights with $W:=\sum_{i=1}^{\infty}w_i<\infty$, and let $\{x_i\}_{i\ge1}$ and $\{y_i\}_{i\ge1}$ be bounded non-negative sequences that are both non-decreasing in $i$. Then
\begin{equation}\label{eq:C.0}
\left(\sum_{i=1}^{\infty}w_ix_i\right)\left(\sum_{i=1}^{\infty}w_iy_i\right)
\le W\sum_{i=1}^{\infty}w_ix_iy_i.
\end{equation}
In particular, under Assumption~\ref{ass:3.1} with $0<\gamma<1/\lambda_1$, the propagation kernel $k(h)=\sum_{i\ge1}\lambda_i^2(1-\gamma\lambda_i)^{2h}$ satisfies $k(u)k(v)\le k(0)k(u+v)=\tr(\Hb^2)k(u+v)$ for all $u,v\ge0$.
\end{lemma}
\begin{proof}[Proof of Lemma~\ref{lem:C.0}]
Since $\{x_i\}$ and $\{y_i\}$ are both non-decreasing, $(x_i-x_j)(y_i-y_j)\ge0$ for all $i,j\ge1$. Multiplying by $w_iw_j\ge0$ and summing over all $i,j\ge1$ gives
\[
0\le\frac12\sum_{i=1}^{\infty}\sum_{j=1}^{\infty}w_iw_j(x_i-x_j)(y_i-y_j)
=W\sum_{i=1}^{\infty}w_ix_iy_i-\left(\sum_{i=1}^{\infty}w_ix_i\right)\left(\sum_{j=1}^{\infty}w_jy_j\right),
\]
which proves \eqref{eq:C.0}. Taking $w_i=\lambda_i^2$, $x_i=(1-\gamma\lambda_i)^{2u}$, and $y_i=(1-\gamma\lambda_i)^{2v}$ (which are non-decreasing in $i$ since $\lambda_1\ge\lambda_2\ge\cdots>0$) yields $k(u)k(v)\le\tr(\Hb^2)k(u+v)$.
\end{proof}

\begin{lemma}[Polynomial Exponent-Halving Sandwich]\label{lem:A.1}
For every $x\in[0,1]$ and integer $N\ge2$ with $m:=\floor{N/2}\ge1$, it holds that $1-x^m\le1-x^N\le3(1-x^m)$. Consequently, for any PSD operator $\Hb$ with $\norm{\gamma\Hb}_2\le1$ and any PSD operator $\Ab\succeq\zero$,
\begin{equation}\label{eq:A.7}
\frac13\tr\Big(\big(\Ib-(\Ib-\gamma\Hb)^{2N}\big)\Ab\Big)
\le\tr\Big(\big(\Ib-(\Ib-\gamma\Hb)^{2\floor{N/2}}\big)\Ab\Big)
\le\tr\Big(\big(\Ib-(\Ib-\gamma\Hb)^{2N}\big)\Ab\Big).
\end{equation}
\end{lemma}
\begin{proof}[Proof of Lemma~\ref{lem:A.1}]
Since $m=\floor{N/2}\ge1$ satisfies $m\le N\le3m$ and $x\in[0,1]$, we have $1-x^m\le1-x^N\le1-x^{3m}=(1-x^m)(1+x^m+x^{2m})\le3(1-x^m)$. Applying this to each eigenvalue $x=(1-\gamma\lambda_i)^2\in[0,1]$ yields~\eqref{eq:A.7}.
\end{proof}

\begin{lemma}[Exact Two-Sided Bound for Finite Geometric Convolutions]\label{lem:A.2}
For any scalars $p,q\in(0,1]$ and any integer $s\ge1$, let $M:=p\vee q$ and $m:=p\wedge q$. Then
\begin{equation}\label{eq:A.8}
(1-e^{-1})M^{s-1}\min\left\{s,\frac{M}{M-m}\right\}
\le\sum_{r=0}^{s-1}p^rq^{s-1-r}
\le M^{s-1}\min\left\{s,\frac{M}{M-m}\right\},
\end{equation}
where $M/(M-m)$ is interpreted as $+\infty$ when $p=q$.
\end{lemma}
\begin{proof}[Proof of Lemma~\ref{lem:A.2}]
By symmetry, assume $p\le q$, so $M=q$, $m=p$, and set $u:=p/q\in(0,1]$. Then $\sum_{r=0}^{s-1}p^rq^{s-1-r}=q^{s-1}\sum_{r=0}^{s-1}u^r$. Since $u\in(0,1]$, we have $\sum_{r=0}^{s-1}u^r\le s\wedge(1-u)^{-1}=\min\{s,q/(q-p)\}$. Conversely, if $1-u\le1/s$, then $u\ge1-1/s$ gives $\sum_{r=0}^{s-1}u^r\ge s(1-(1-1/s)^s)\ge(1-e^{-1})s$; if $1-u>1/s$, then $u^s<(1-1/s)^s<e^{-1}$ gives $\sum_{r=0}^{s-1}u^r=(1-u^s)/(1-u)\ge(1-e^{-1})q/(q-p)$.
\end{proof}

\subsection{Operator Lemmas from Prior Work}\label{app:C.2}

\begin{lemma}[Sub-Gaussian Implies Fourth-Moment Upper Bound; \citealt{zou2021}, Lemma~A.1; \citealt{hsu2014}; \citealt{vershynin2018}, Proposition~2.5.2]\label{lem:C.1}
Suppose $\EE[\xb\xb^\top]=\Hb\succ\zero$ and the standardized feature projections are $\sigma_z^2$-sub-Gaussian, in the sense that $\EE[\exp(\la\ub,\xb\ra)]\le\exp(\sigma_z^2\norm{\ub}_{\Hb}^2/2)$ for all $\ub\in\cH$ (equivalently, $\EE[\exp(\lambda\la\vb,\Hb^{-1/2}\Pb_m\xb\ra)]\le\exp(\lambda^2\sigma_z^2/2)$ for all finite-rank spectral projections $\Pb_m=\Ib_{0:m}$, unit vectors $\vb\in\cH$, and $\lambda\in\RR$). Then for any PSD operator $\Ab\succeq\zero$, Assumption~\ref{ass:3.2} holds with $\alpha=16\sigma_z^4$:
\[
\cM\circ\Ab=\EE\bigl[(\xb^\top\Ab\xb)\xb\xb^\top\bigr]\preceq16\sigma_z^4\tr(\Hb\Ab)\Hb.
\]
\end{lemma}

\begin{lemma}[Properties of PSD Super-Operators; \citealt{zou2021}, Lemmas~4.1 and~B.1; \citealt{jain2017a,jain2017b}]\label{lem:C.2}
A linear super-operator $\cO$ on symmetric operators is called a PSD mapping if $\Ab\succeq\zero$ implies $\cO\circ\Ab\succeq\zero$. Under Assumptions~\ref{ass:3.1} and \ref{ass:3.2}, the super-operators defined in \eqref{eq:3.3} and \eqref{eq:5.3} are well-defined on symmetric trace-class operators and satisfy:
\begin{enumerate}
\item $\cM$ and $\wt\cM$ are both PSD mappings.
\item $\cI-\gamma\cT$ and $\cI-\gamma\wt\cT$ are both PSD mappings for all $\gamma>0$.
\item $\cM-\wt\cM$ and $\wt\cT-\cT=\gamma(\cM-\wt\cM)$ are both PSD mappings; in particular, $(\cI-\gamma\cT)\circ\Ab\succeq(\cI-\gamma\wt\cT)\circ\Ab=(\Ib-\gamma\Hb)\Ab(\Ib-\gamma\Hb)$ for all $\Ab\succeq\zero$.
\end{enumerate}
\end{lemma}

\begin{lemma}[Bias-Variance Risk Decomposition; \citealt{zou2021}, Lemmas~B.2, C.1, D.1, and~D.6]\label{lem:C.3}
Consider SGD with tail-averaging \eqref{eq:3.2} and recall the definitions of $\bias$ and $\variance$ in \eqref{eq:5.6}. Under Assumptions~\ref{ass:3.1}, \ref{ass:3.2}, and \ref{ass:3.3} (which ensure $\EE\norm{\bmeta_t^{\bias}}_{\Hb}^2<\infty$ and $\EE\norm{\bmeta_t^{\variance}}_{\Hb}^2<\infty$ for all $t\ge0$):
\begin{enumerate}
\item For any data distribution $\cD$, the expected excess risk satisfies
\begin{align*}
\EE\bigl[L(\overbar{\wb}_{s:s+N})\bigr]-L(\wb^*)
&=\frac12\inner{\Hb}{\EE\bigl[\overbar{\bmeta}_{s:s+N}\otimes\overbar{\bmeta}_{s:s+N}\bigr]}\\
&\le\bigl(\sqrt{\bias}+\sqrt{\variance}\bigr)^2\le2\cdot\bias+2\cdot\variance.
\end{align*}
\item If the residual noise $\xi_t=y_t-\la\wb^*,\xb_t\ra$ is well-specified as in \eqref{eq:3.4} (so $\EE[\xi_t\mid\xb_t]=0$), then $\EE[\overbar{\bmeta}_{s:s+N}^{\bias}\otimes\overbar{\bmeta}_{s:s+N}^{\variance}]=\zero$ and the excess risk decomposes with exact equality:
\[
\EE\bigl[L(\overbar{\wb}_{s:s+N})\bigr]-L(\wb^*)=\bias+\variance.
\]
\end{enumerate}
\end{lemma}

\begin{lemma}[Single-Step Trace Identity and Interval Telescoping Inequality; adapted from \citealt{zou2021}, Eqs.~(B.1) and~(D.4)]\label{lem:C.4}
Under Assumptions~\ref{ass:3.1} and \ref{ass:3.2}, if $\gamma<1/(\alpha\tr(\Hb))$, then for any integers $0\le a<b$ and $\Bb_{a,b}:=\Bb_a-(\Ib-\gamma\Hb)^{b-a}\Bb_a(\Ib-\gamma\Hb)^{b-a}$, the bias covariance sequence $\{\Bb_u\}_{u\ge0}$ satisfies
\begin{equation}\label{eq:C.4}
\gamma\sum_{u=a}^{b-1}\tr(\Hb\Bb_u)
\le\frac{\tr(\Bb_a)-\tr(\Bb_b)}{2-\gamma\alpha\tr(\Hb)}
\le\frac1{2\bigl(1-\gamma\alpha\tr(\Hb)\bigr)}\tr(\Bb_{a,b}).
\end{equation}
\end{lemma}
\begin{proof}[Proof of Lemma~\ref{lem:C.4}]
We adapt the telescoping argument of \citet[Eqs.~(B.1) and~(D.4)]{zou2021} to a general interval $[a,b]$ for completeness. For each step $u\ge0$, taking the trace of $\Bb_{u+1}=(\cI-\gamma\cT)\circ\Bb_u=\Bb_u-\gamma(\Hb\Bb_u+\Bb_u\Hb)+\gamma^2\cM\circ\Bb_u$ and applying Assumption~\ref{ass:3.2} ($\tr(\cM\circ\Bb_u)\le\alpha\tr(\Hb)\tr(\Hb\Bb_u)$) gives
\[
\tr(\Bb_{u+1})\le\tr(\Bb_u)-\gamma\bigl(2-\gamma\alpha\tr(\Hb)\bigr)\tr(\Hb\Bb_u).
\]
Summing this inequality over $u\in\{a,\ldots,b-1\}$ and dividing by $2-\gamma\alpha\tr(\Hb)>2(1-\gamma\alpha\tr(\Hb))>0$ yields the first inequality in \eqref{eq:C.4}. Finally, since $\Bb_b=(\cI-\gamma\cT)^{b-a}\circ\Bb_a\succeq(\Ib-\gamma\Hb)^{b-a}\Bb_a(\Ib-\gamma\Hb)^{b-a}$ by Lemma~\ref{lem:C.2}(3), we have $\tr(\Bb_a)-\tr(\Bb_b)\le\tr(\Bb_{a,b})$, establishing the second inequality in \eqref{eq:C.4}.
\end{proof}

\begin{lemma}[Upper Bounds on Constant-Stepsize Variance Iterates; \citealt{zou2021}, Lemmas~B.4 and~B.5]\label{lem:C.5}
Under Assumptions~\ref{ass:3.1}, \ref{ass:3.2}, and \ref{ass:3.3}, if the constant stepsize satisfies $0<\gamma<1/(\alpha\tr(\Hb))$, then the variance covariance sequence $\{\Cb_t\}_{t\ge0}$ defined in \eqref{eq:5.5} satisfies
\[
\Cb_t\preceq\frac{\gamma\sigma^2}{1-\gamma\alpha\tr(\Hb)}\bigl(\Ib-(\Ib-\gamma\Hb)^t\bigr)
\preceq\frac{\gamma\sigma^2}{1-\gamma\alpha\tr(\Hb)}\Ib,
\qquad\forall t\ge0.
\]
\end{lemma}

\clearpage

\bibliographystyle{ims}
\bibliography{references}
\end{document}